%% file: main.tex
\documentclass[11pt]{article}

\usepackage[preprint]{acl}
\usepackage{times}
\usepackage{latexsym}

\usepackage{algorithm}
\usepackage{algpseudocode}

\usepackage{amsmath,amssymb,amsthm}
\usepackage{bbm}

\usepackage{graphicx}
\usepackage{subcaption}
\usepackage{booktabs}
\usepackage{multirow,multicol}
\usepackage{stfloats}

\usepackage{tcolorbox}
\usepackage{enumitem}
\usepackage{pifont}
\usepackage{xcolor}

\usepackage[T1]{fontenc}
\usepackage[utf8]{inputenc}
\usepackage{microtype}
\usepackage{inconsolata}

\newcommand{\vect}[1]{\mathbf{#1}}
\newcommand{\mat}[1]{\mathbf{#1}}
\newtheorem{theorem}{Theorem}

\title{{\normalfont\small Accepted at the 2026 Conference on Empirical Methods in Natural Language Processing (EMNLP 2026)\par}\vspace{0.7em}
Drift No More: Spectral Locking for Long-Horizon Open-Weight Multi-Agent AI}

\author{
  \textbf{Vishal Pramanik\textsuperscript{1}},
  \textbf{Maisha Maliha\textsuperscript{2}},
  \textbf{Susmit Jha\textsuperscript{3}},
  \textbf{Sumit Kumar Jha\textsuperscript{1}}
\\
\\
  \textsuperscript{1}University of Florida,
  \textsuperscript{2}University of Oklahoma,
  \textsuperscript{3}SRI International
\\
  \small{
    \texttt{vishalpramanik@ufl.edu},
    \texttt{maisha.maliha-1@ou.edu},
    \texttt{susmit.jha@berkeley.edu},
    \texttt{sumit.jha@ufl.edu}
  }
}

\begin{document}
\maketitle

\begin{abstract}
Long-horizon multi-agent systems suffer from three tightly coupled failures: task drift, role drift, and state staleness. We test the hypothesis that these share a common representational cause: agent hidden states move away from the directions encoding the active task, assigned role, and current shared state. A pilot study over 14{,}832 annotated turns supports this: drifted turns collapse near-uniformly across all three anchor subspaces (pairwise $r\!=\!0.72$--$0.78$). Existing methods address these failures separately, operating at the surface-text level or treating control directions independently. We introduce \textbf{ATLAS} (\textbf{A}lgebraic \textbf{T}ask--role \textbf{L}ocking with \textbf{A}daptive \textbf{S}pectral Memory), a training-free intervention acting on hidden states before a token is generated. ATLAS builds a positive-semidefinite anchor kernel over task, role, and state subspaces and steers hidden states toward their shared high-coherence eigendirections; a low-rank Gram decomposition keeps this tractable. Across long-horizon memory, state maintenance, tool use, least-privilege coordination, and embodied stress tests, ATLAS outperforms prompting, retrieval, memory, and activation-steering baselines. Against A-MEM it improves memory macro-average accuracy by up to 5.7 points and reduces write/read failures by up to 31.7\%/24.8\%; against MemoryArena-Agent it gains up to 4.8 points. It gains 10.7--12.1 points under perturbation, reaches 93.5\% role consistency, and transfers to a 72B backbone (macro-average 61.0 vs.\ 56.2). Ablating the joint task--role--state span costs 7.2 points against 0.8 for a matched-rank random span, indicating these directions are functionally implicated, not incidental; supplying them by prompt or retrieval recovers $4.3$ points less. ATLAS is therefore a representation-level controller for open-weight multi-agent systems, complementary to output-level protocols for black-box agents.\footnote{Code: \url{https://github.com/VishalPramanik/ATLAS}}
\end{abstract}

\input{Sections/Introduction}
\input{Sections/Method}
\input{Sections/Experiment}
\input{Sections/Ablation}

\bibliography{custom}
\clearpage
\input{Sections/Appendix_final}

\end{document}

%% file: Sections/Introduction.tex
\begin{figure}[t]
  \centering
  \raisebox{-0.0pt}{\includegraphics[width=219.65521pt]{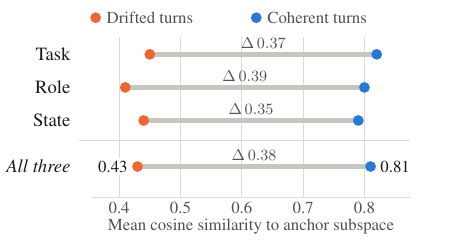}}
  \caption{Anchor alignment at layer $\ell^\star$ over 14{,}832
  annotated turns (200 ten-agent Qwen3-8B conversations). Drifted
  turns collapse from ${\sim}0.80$ to ${\sim}0.43$, near-uniformly
  across all three anchors.}
  \label{fig:pilot_alignment}
\end{figure}

\begin{figure*}[t]
    \centering
    \raisebox{-0.0pt}{\includegraphics[width=451.71432pt]{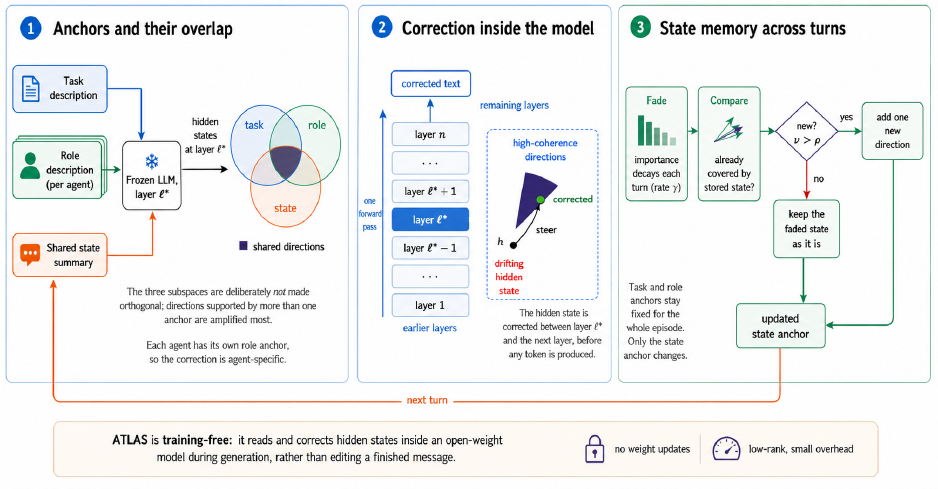}}
    \caption{Overview of ATLAS. ATLAS constructs task, role, and state anchor
subspaces at layer $\ell^\star$, combines them into an agent-specific low-rank
PSD kernel $F^{(t)}(F^{(t)})^\top$, and steers hidden states toward coherent
high-eigenvalue directions. The state component is updated by uniformly
decaying existing state eigenvalues and appending a novel residual direction
when $\nu>\rho$, after which the corrected hidden state is passed to downstream
layers.}
    \label{fig:atlas_overview}
\end{figure*}

\section{Introduction}
\label{sec:introduction}

Large language model agents are increasingly deployed as collaborative systems in which multiple specialized agents plan, retrieve, write memory, call tools, verify outputs, and coordinate over long interaction traces~\citep{li2023camel,wu2023autogen,park2023generativeagents}. This decomposition is useful when a task requires persistent memory, role specialization, and tool-mediated execution rather than a single monolithic response. It also creates a failure mode that ordinary single-turn accuracy does not capture: as an interaction lengthens, an agent may continue to produce fluent, locally plausible text while its internal representation moves away from the active task, its assigned role, or the latest shared state. We refer to these three coupled failures as \emph{task drift}, \emph{role drift}, and \emph{state staleness}. Recent memory and agent benchmarks show that long-term conversational memory, on-policy state maintenance, and multi-session memory use remain difficult even for strong LLM systems~\citep{hu2026evermembench,jiayang2026amemgym,he2026memoryarena}. In a long-horizon multi-agent workflow, a single stale write, role-inappropriate assertion, or unauthorized tool action can propagate across many downstream turns.

Existing approaches address parts of this problem, but at different control surfaces. Role prompts and structured-agent frameworks specify agent responsibilities, yet a prompt-level schema does not itself maintain the model's internal task--role--state representation over a long trajectory~\citep{li2023camel}. Retrieval-augmented and agentic-memory systems improve access to past information, but they do not directly correct a hidden state that is already using stale or role-inappropriate information~\citep{park2023generativeagents,xu2025amem}. Scoped-context methods enforce access boundaries rather than repairing representational drift~\citep{agostino2026alara}. Activation-steering methods do intervene inside the model, but commonly construct control directions independently, whereas task, role, and state constraints in collaborative work are coupled~\citep{turner2024actadd}. Each therefore improves one isolated aspect of coordination---prompt discipline, retrieval, memory organization, or activation control---without jointly stabilizing task, role, and state. The question addressed here is specifically \emph{representation-level}: when intermediate model states are accessible, can a single intervention preserve the shared geometry of task, role, and evolving state over a long interaction?

We introduce \textbf{ATLAS}, an anchor-kernel intervention for this setting. For each agent, ATLAS represents the active task, assigned role, and current shared state as subspaces at an intervention layer, and combines them into a single positive-semidefinite kernel that retains their overlap. Hidden states are then steered toward directions that are jointly supported by these anchors, and the state component is refreshed as the conversation evolves. This places ATLAS at a different control surface from output-level role enforcement, which inspects or revises a message at the agent boundary and therefore needs no access to internal activations: ATLAS instead requires hidden-state access and corrects the representation \emph{during} generation, before a token is emitted.

Our contributions are threefold. First, we formulate task drift, role drift, and state staleness as coupled representational failures in long-horizon multi-agent systems, and support this with a pilot study over 14{,}832 annotated turns. Second, we develop a low-rank spectral anchor kernel with adaptive state refresh for inference-time hidden-state correction. Third, we evaluate the method across homogeneous, expanded, and heterogeneous teams on long-horizon memory, interactive state maintenance, tool use, least-privilege coordination, and appendix-only embodied stress tests, and probe the functional role of the learned anchor geometry through targeted subspace removal. The resulting evidence establishes when a representation-level controller helps, and where its access assumptions limit deployment.

\section{Motivation}
\label{sec:motivation}

We now state the hypothesis that organizes the rest of the paper. Task drift, role drift, and state staleness are not independent failures but surface manifestations of a single representational one: the agent's hidden state moves away from the directions encoding the current task, assigned role, and latest shared state. We evaluate this hypothesis below with diagnostic, temporal, and interventional evidence. If it holds, one intervention that restores alignment along all three directions should outperform methods that target each failure mode independently.

We test this hypothesis in a pilot study of 200 ten-agent Qwen3-8B
conversations comprising 14{,}832 turns labeled as coherent or drifted using
a frontier LLM-as-judge protocol with GPT-5.5 as the primary evaluator; the
full protocol is provided in Appendix~\ref{app:pilot_study}. For each turn, we
measure the cosine similarity between the agent's hidden state at layer
$\ell^\star$ and the task, role, and state anchor subspaces.
Figure~\ref{fig:pilot_alignment} shows that coherent turns remain strongly
aligned with all three subspaces, whereas drifted turns undergo a near-uniform
collapse across task, role, and state. The failures are also tightly coupled,
with pairwise Pearson correlations of $r=0.78$ (task--role), $0.74$
(task--state), and $0.72$ (role--state), indicating that the onset of one drift
mode reliably predicts the others.

This coupling explains why existing approaches recover so little of the lost
alignment (Appendix Table~\ref{tab:pilot_full}). Prompt-level schemas and retrieval
mechanisms recover at most $+0.03$ over Prompt-Only, because they modify inputs
without correcting internal geometry. Activation-steering methods reach
$+0.06$, but their independent or orthogonalized directions discard the
task--role--state overlap that characterizes coordination failure. ATLAS
recovers $+0.31$---more than five times the strongest alternative---by encoding
all three anchors in a single positive-semidefinite kernel whose shared
eigendirections concentrate correction where the failure modes are most
entangled. The remainder of the paper formalizes this mechanism
and evaluates it across benchmarks spanning long-horizon memory, tool use,
least-privilege coordination, and embodied stress tests.

%% file: Sections/Method.tex
\section{Method}
\label{sec:method}

\subsection{Overview}
\label{sec:method_overview}

ATLAS operates on intermediate representations, not on completed agent messages. At turn $t$, the method asks whether the hidden state of agent $a$ remains jointly compatible with three pieces of information: the active task, the authority attached to the agent's role, and the current shared state. These are encoded as anchor subspaces and combined into a single spectral object, the \emph{anchor kernel}. The kernel is deliberately non-orthogonal: directions supported by more than one anchor should exert greater influence, because task, role, and state are not independent in collaborative work.

For token $i$ of agent $a$ at turn $t$, let $\vect{h}^{(a,i)}_t \in \mathbb{R}^d$ denote the hidden state at intervention layer $\ell^\star$. ATLAS constructs task, role, and state anchor subspaces from anchor texts, combines them in a positive-semidefinite kernel $\mat{K}^{(t)}$, and steers $\vect{h}^{(a,i)}_t$ toward the high-eigenvalue directions of that kernel before the next layer is evaluated. After each turn, only the state component is refreshed. The mechanism therefore carries a clear access assumption: it requires a model whose hidden states can be read and modified at inference time, unlike message-boundary controllers that detect or repair a completed output without intervening in the generator's latent state.

We select $\ell^\star$ in the middle third of the model using a linear probe that separates on-topic from off-topic continuations, following prior representation-engineering work~\citep{zou2023representation,li2023iti}. Appendix~\ref{app:probe} gives the layer-selection details. Corrections are applied after layer $\ell^\star$ and before layer $\ell^\star+1$, so downstream layers receive the corrected representation.

\subsection{Anchor Subspaces}
\label{sec:anchor_subspaces}

ATLAS begins by converting task, role, and state descriptions into geometric anchors. For each anchor type
\[
c \in \{T, R^{(a)}, S^{(t)}\},
\]
we pass the corresponding anchor text through the frozen model and collect hidden states at layer $\ell^\star$. These hidden states are stacked into a matrix
\[
\mat{X}_c \in \mathbb{R}^{d \times m_c}.
\]
A thin SVD gives
\[
\mat{X}_c
=
\mat{U}_c \mat{S}_c \mat{W}_c^\top .
\]
We retain the smallest number of directions $k_c$ that explain a variance fraction $\tau$:
\begin{equation}
\label{eq:basis_main}
k_c =
\min
\left\{
k :
\frac{\sum_{j=1}^{k} s_j^2}
     {\sum_j s_j^2}
\ge \tau
\right\}.
\end{equation}
The resulting basis $\mat{U}_c \in \mathbb{R}^{d \times k_c}$ represents the subspace for that anchor. In our experiments, $k_c$ is typically between 5 and 30 for hidden dimension $d=4096$.

ATLAS therefore does \emph{not} orthogonalize these bases. The entanglement is concrete: a Planner's next step is task-specific, constrained by its role, and conditioned on the latest memory, so a single hidden direction can carry all three constraints at once. Orthogonalizing discards exactly that shared structure, and costs $3.8$ points on EverMemBench ($50.3$ vs.\ $54.1$; Appendix~\ref{app:overlap}).

\subsection{Anchor Kernel}
\label{sec:kernel}

The anchor kernel combines the three anchor sources into one positive-semidefinite matrix. For agent $a$ at turn $t$, we define
\begin{equation}
\label{eq:kernel_main}
\begin{aligned}
\mat{K}^{(t)}
&=
w_T \mat{U}_T \mat{U}_T^\top
+
w_R \mat{U}^{(a)}_R
       (\mat{U}^{(a)}_R)^\top      \\
&\quad+
w_S \mat{V}^{(t)}_S
       \mat{\Sigma}^{(t)}_S
       (\mat{V}^{(t)}_S)^\top ,
\end{aligned}
\end{equation}
where $w_T+w_R+w_S=1$. The first two terms encode the fixed task and role anchors. The third term is the evolving state component. Here, $\mat{V}^{(t)}_S$ contains state directions and $\mat{\Sigma}^{(t)}_S$ stores their importance values.

The kernel gives stronger weight to directions supported by multiple anchors. If a direction is relevant to both the task and the current state, its effective eigenvalue receives contributions from both terms. This is the main reason we use a joint PSD kernel instead of independently steering along task, role, and state vectors.

Although $\mat{K}^{(t)}$ is a $d \times d$ matrix, ATLAS never needs to form it explicitly. We use the low-rank factorization
\begin{equation}
\label{eq:factor_main}
\mat{K}^{(t)}
=
\mat{F}^{(t)}
(\mat{F}^{(t)})^\top ,
\end{equation}
where
\begin{equation}
\label{eq:F_main}
\begin{aligned}
\mat{F}^{(t)}
=
\big[
&\sqrt{w_T}\mat{U}_T,\;
 \sqrt{w_R}\mat{U}^{(a)}_R, \\
&\sqrt{w_S}\mat{V}^{(t)}_S
  (\mat{\Sigma}^{(t)}_S)^{1/2}
\big].
\end{aligned}
\end{equation}
If $p=k_T+k_R+k_S$, then $\mat{F}^{(t)} \in \mathbb{R}^{d \times p}$ with $p \ll d$. We compute eigenpairs using the small Gram matrix
\[
\mat{G}^{(t)}
=
(\mat{F}^{(t)})^\top \mat{F}^{(t)}
\in \mathbb{R}^{p \times p}.
\]
The nonzero eigenvalues of $\mat{G}^{(t)}$ and $\mat{K}^{(t)}$ are identical, and the eigenvectors of $\mat{K}^{(t)}$ are recovered by lifting. This reduces eigendecomposition cost from cubic in $d$ to $O(dp^2+p^3)$. Full details are in Appendix~\ref{app:kernel_computation}.

\subsection{Coherence Score}
\label{sec:score}

Given the anchor kernel, ATLAS measures whether a hidden state is aligned with the current task--role--state geometry. Let $\mat{K}^{(t)}=\sum_{j=1}^{r}\lambda_j \vect{v}_j \vect{v}_j^\top$ be the nonzero eigendecomposition of the kernel. The coherence of a hidden state $\vect{h}$ is its Rayleigh quotient:
\begin{equation}
\label{eq:coherence_main}
\mathcal{C}(\vect{h})
=
\frac{\vect{h}^\top
      \mat{K}^{(t)}
      \vect{h}}
     {\|\vect{h}\|^2}.
\end{equation}
Equivalently,
\begin{equation}
\label{eq:coherence_expanded_main}
\mathcal{C}(\vect{h})
=
\sum_{j=1}^{r}
\lambda_j p_j(\vect{h}),
\qquad
p_j(\vect{h})
=
\frac{
(\vect{v}_j^\top \vect{h})^2}
{\|\vect{h}\|^2}.
\end{equation}
Thus, $\mathcal{C}(\vect{h})$ is high when the hidden state places more mass on high-eigenvalue anchor directions.

We also use the spectral variance
\begin{equation}
\label{eq:spectral_variance_main}
\mathcal{V}(\vect{h})
=
\frac{
\vect{h}^\top
(\mat{K}^{(t)})^2
\vect{h}}
{\|\vect{h}\|^2}
-
\mathcal{C}(\vect{h})^2 .
\end{equation}
This quantity is small when $\vect{h}$ is already concentrated in a coherent eigenspace, and larger when its mass is spread across inconsistent eigendirections. Appendix~\ref{app:steering_analysis} shows that this variance controls the coherence increase induced by ATLAS steering.

\subsection{Spectral Steering}
\label{sec:steer}

ATLAS steers each token by amplifying the kernel-aligned components of its hidden state:
\begin{equation}
\label{eq:steer_main}
\widetilde{\vect{h}}^{(a,i)}_t
=
(\mat{I}
+
\alpha_{\mathrm{eff}}\mat{K}^{(t)})
\vect{h}^{(a,i)}_t .
\end{equation}
Using the eigendecomposition of $\mat{K}^{(t)}$, this can be written as
\begin{equation}
\label{eq:steer_expanded_main}
\begin{aligned}
\widetilde{\vect{h}}^{(a,i)}_t
&=
\vect{h}^{(a,i)}_t  \\
&\quad+
\alpha_{\mathrm{eff}}
\sum_{j=1}^{r}
\lambda_j
(\vect{v}_j^\top \vect{h}^{(a,i)}_t)
\vect{v}_j .
\end{aligned}
\end{equation}
The gain is capped as
\begin{equation}
\label{eq:alpha_eff_main}
\alpha_{\mathrm{eff}}
=
\min(\alpha,\xi/\lambda_1),
\end{equation}
where $\xi$ limits the largest per-direction correction. We use $\xi=0.15$ by default.

Under the continuous dynamics $d\vect{h}/d\alpha=\mat{K}\vect{h}$, coherence is
non-decreasing, with $d\mathcal{C}/d\alpha=2\mathcal{V}(\vect{h})\geq0$. The
increase is larger when $\vect{h}$ has mass across eigenspaces with different
eigenvalues, but can vanish when it is concentrated in a single eigenspace,
including the null space. Appendix~\ref{app:steering_analysis} proves this
continuous result and the monotonicity of the discrete steering update in
Eq.~\eqref{eq:steer_main}.

\subsection{Adaptive State Memory}
\label{sec:memory}

The task and role anchors remain fixed during an episode, but the state anchor must evolve as the conversation changes. ATLAS updates only the state component $\mat{K}^{(t)}_{\mathrm{state}}$ after each turn. The update has three steps: decay, novelty extraction, and enrichment.

First, existing state eigenvalues are decayed:
\begin{equation}
\label{eq:decay_main}
\bar{\sigma}^{(t)}_j
=
\gamma \sigma^{(t)}_j,
\qquad
j=1,\ldots,k_S .
\end{equation}
This geometrically reduces the influence of all previously stored state
directions. Novel residual directions are appended during the enrichment step
described below. Second, ATLAS computes a turn-level representation from the steered token states:
\begin{equation}
\label{eq:turn_mean_main}
\bar{\widetilde{\vect{h}}}^{(a)}_t
=
\frac{1}{L_t}
\sum_{i=1}^{L_t}
\widetilde{\vect{h}}^{(a,i)}_t .
\end{equation}
It then projects this representation away from the current state subspace:
\begin{equation}
\label{eq:novel_direction_main}
\vect{m}^{(a)}_t
=
\left(
\mat{I}
-
\mat{V}^{(t)}_S
(\mat{V}^{(t)}_S)^\top
\right)
\bar{\widetilde{\vect{h}}}^{(a)}_t .
\end{equation}
The novelty score is
\begin{equation}
\label{eq:novelty_main}
\nu
=
\frac{
\|\vect{m}^{(a)}_t\|^2}
{\|\bar{\widetilde{\vect{h}}}^{(a)}_t\|^2}.
\end{equation}
If $\nu \le \rho$, the turn is treated as a restatement of existing state. If $\nu>\rho$, the normalized novel direction is appended to the state basis:
\begin{equation}
\label{eq:append_state_main}
\begin{aligned}
\mat{V}^{(t+1)}_S
&\leftarrow
[\mat{V}^{(t)}_S,\; \hat{\vect{m}}^{(a)}_t], \\[2pt]
\hat{\vect{m}}^{(a)}_t
&=
\frac{\vect{m}^{(a)}_t}{\|\vect{m}^{(a)}_t\|}.
\end{aligned}
\end{equation}
The new direction receives initial importance $\beta_0$:
\begin{equation}
\label{eq:update_sigma_main}
\mat{\Sigma}^{(t+1)}_S
\leftarrow
\mathrm{diag}
(
\bar{\sigma}^{(t)}_1,\ldots,
\bar{\sigma}^{(t)}_{k_S},
\beta_0
).
\end{equation}
If the state rank exceeds $k_S^{\max}$, ATLAS drops the smallest-eigenvalue state direction. Every $\Delta$ turns, the state basis is re-orthonormalized. This update lets ATLAS forget stale information while incorporating new decisions, constraints, and corrections. Appendix~\ref{app:memory_update} gives implementation details and Appendix~\ref{app:state_stability} gives the stability bound for enrichment. Figure~\ref{fig:atlas_overview} summarizes the intervention, and Algorithm~\ref{alg:atlas} gives the complete procedure.

%% file: Sections/Experiment.tex

\section{Experiments and Results}
\label{sec:experiments}

\subsection{Setup}
\label{ssec:exp_setup}

We evaluate the coupled-failure hypothesis of \S\ref{sec:motivation} under three increasingly difficult multi-agent settings. \textbf{T1} is a homogeneous ten-agent role-structured team. It follows the same basic role decomposition used by role-prompting baselines such as Struct-Role, but all agents share one backbone so that the setting isolates coordination under a common representation space. \textbf{T2} is a homogeneous twelve-agent expanded memory team. It is based on the memory-centric design of \mbox{A-MEM}~\cite{xu2025amem}, but we increase the number of agents and split memory, verification, planning, and summarization into more specialized roles to increase coordination complexity while keeping the backbone fixed. \textbf{T3} is a heterogeneous twelve-agent team in which the twelve roles are distributed across six different model families. This is the hardest setting: agents differ not only in role and context, but also in capability profile and latent representation geometry. The ten-agent set spans coordination, planning, memory read/write, execution, verification, critique, routing, summarization, and monitoring; T2 and T3 add Tool-Executor, the only role permitted to invoke external tools, and Safety-Checker, which validates privilege boundaries, unsafe actions, and out-of-scope context exposure. Full role definitions, privilege boundaries, and agent-to-model assignments are in Appendix~\ref{app:agent_arch} and~\ref{app:backbones}.

\paragraph{Backbones and heterogeneous team.}
For T1 and T2, all agents within a run share the same backbone. T1 uses Llama-4-Scout-17B-16E~\cite{meta2025llama4scout}, and T2 uses Qwen3-8B~\cite{qwen2025qwen3}, which we also use as the primary backbone for ablations because it provides a strong trade-off between capability and evaluation cost. For T3, the twelve agents are distributed across six model families: Our heterogeneous agent stack comprises Llama-4-Scout-17B-16E-Instruct~\cite{meta2025llama4scout}, Qwen3-8B~\cite{qwen2025qwen3}, Qwen2.5-7B-Instruct~\cite{qwen2024qwen25}, Phi-4 (14B)~\cite{microsoft2024phi4}, Gemma-2-9B-IT~\cite{gemmateam2024gemma2}, and Mistral-7B-Instruct-v0.1~\cite{jiang2023mistral7b}. The heterogeneous assignment places stronger reasoning models on planning and routing, compact instruction-following models on memory operations, and separate critic-style models on verification, monitoring, and safety checking. Exact assignments are listed in Appendix~\ref{app:backbones}.

\paragraph{Benchmarks and protocols.}
The main paper focuses on two primary text-based long-horizon coordination benchmarks:
\begin{enumerate}[leftmargin=*,nosep]
\item \textbf{EverMemBench}~\cite{hu2026evermembench}: 2{,}400 QA pairs over long multi-party conversations, evaluating long-horizon memory across Fine-Grained Recall (FGR), Memory Awareness (MA), and Profile Understanding (PU).
\item \textbf{AMemGym}~\cite{jiayang2026amemgym}: 1{,}000 interactive, on-policy episodes per configuration, where earlier system choices influence later observations, with diagnostic write/read/utilization failure rates.
\end{enumerate}
We report two secondary text-based stress tests in the appendix: ToolACE~\cite{liu2025toolace}, which probes robustness in multi-step tool-use chains, and ALARA~\cite{agostino2026alara}, which evaluates least-privilege context discipline in agentic workflows. We also include LIBERO~\cite{liu2023libero} as an appendix-only multimodal embodied stress test. Because LIBERO is not a native multi-agent dialogue benchmark, we wrap each language-conditioned manipulation episode in our multimodal multi-agent controller and inject one of three controlled perturbations: \emph{role-confuse}, \emph{state-delay}, or \emph{task-switch}. Full ToolACE and ALARA results are reported in Appendix~\ref{app:secondary_results}, and the complete LIBERO protocol and results are reported in Appendix~\ref{app:libero_results}. Across all runs, we use a 128K token context budget, top-$k\!=\!10$ retrieval, summaries every 20 turns, a 200-turn episode cap, temperature 0.0, and means over three seeds (42, 137, 256) with 95\% paired bootstrap confidence intervals. Full benchmark protocols, hyperparameter search ranges, and reporting conventions are provided in Appendix~\ref{app:exp_protocols}.

\paragraph{Baselines.}
We compare against seven general baselines plus two benchmark-specific recent baselines. \textbf{Single-Agent} uses one monolithic agent with the full context budget, establishing a no-coordination lower bound. \textbf{Prompt-Only} deploys the same agent topology without structured memory or communication control, isolating the effect of agent count alone. \textbf{Struct-Role} adds explicit JSON-schema role declarations per message, representing a role-structured prompting baseline. \textbf{RAG-Mem} equips the topology with a shared vector store and periodic summaries, representing a standard retrieval-augmented memory baseline. \textbf{A-MEM} adapts A-Mem's~\cite{xu2025amem} Zettelkasten-inspired agentic memory---dynamic indexing, linking, and memory evolution---to the multi-agent setting, serving as the strongest memory baseline. \textbf{ActAdd}~\cite{turner2024actadd} applies Activation Addition steering to the Coordinator, Executor, and Monitor, using role-specific contrast pairs to compute steering vectors. \textbf{SEA}~\cite{qiu2024sea} applies Spectral Editing of Activations to the same three agents, representing a prior activation-steering approach.

For memory-specific comparisons, we include \textbf{MemoryArena-Agent}~\cite{he2026memoryarena}, which adapts the MemoryArena memory-agent-environment loop to our EverMemBench and AMemGym settings using the same topology, context budget, retrieval depth, and reporting metrics. We include this baseline only for memory and state-maintenance evaluations, where the comparison is metric-compatible. For least-privilege evaluation, we include \textbf{ALARA-Scoped}~\cite{agostino2026alara}, which uses ALARA's declarative context-agent-tool scoping to enforce per-agent context and tool access. We include this baseline only on ALARA, where the task success, privilege violation, and overexposure metrics directly match its design objective. Together, these baselines test whether ATLAS's gains arise from role prompting, retrieval, memory management, activation steering, scoped context access, or simply the multi-agent topology.

\subsection{Long-Horizon Memory}
\label{ssec:evermem_results}

\begin{table*}[t]
\centering
\caption{\textbf{EverMemBench results} (\%). FGR\,=\,Fine-Grained Recall (\%$\uparrow$), MA\,=\,Memory Awareness (\%$\uparrow$), PU\,=\,Profile Understanding (\%$\uparrow$), Avg\,=\,macro-average (\%$\uparrow$). T1 is a homogeneous 10-agent role-structured team; T2 is a homogeneous 12-agent \mbox{A-MEM}-expanded team; T3 is a heterogeneous 12-agent team using six model families. Values are means over 3 seeds; average standard deviation across cells within each setting block is shown in the header. Best in \textbf{bold}.}
\label{tab:evermem}
\small
\setlength{\tabcolsep}{3.0pt}
\begin{tabular}{@{}l cccc @{\hspace{7pt}} cccc @{\hspace{7pt}} cccc@{}}
\toprule
& \multicolumn{4}{c}{\textbf{T1: Homogeneous-10} {\scriptsize($\bar\sigma\!=\!1.1$)}}
& \multicolumn{4}{c}{\textbf{T2: Homogeneous-12} {\scriptsize($\bar\sigma\!=\!0.8$)}}
& \multicolumn{4}{c}{\textbf{T3: Heterogeneous-12/6M} {\scriptsize($\bar\sigma\!=\!1.3$)}} \\
\cmidrule(lr){2-5}\cmidrule(lr){6-9}\cmidrule(lr){10-13}
& FGR & MA & PU & Avg & FGR & MA & PU & Avg & FGR & MA & PU & Avg \\
\midrule
Single-Agent        & 41.2 & 33.8 & 29.4 & 34.8 & 43.5 & 35.1 & 31.2 & 36.6 & 38.7 & 31.6 & 27.8 & 32.7 \\
Prompt-Only         & 46.3 & 38.2 & 34.5 & 39.7 & 48.1 & 39.7 & 36.0 & 41.3 & 43.8 & 35.9 & 32.1 & 37.3 \\
Struct-Role         & 49.1 & 40.6 & 36.8 & 42.2 & 51.4 & 42.3 & 38.4 & 44.0 & 46.5 & 38.4 & 34.6 & 39.8 \\
RAG-Mem             & 52.7 & 43.1 & 38.9 & 44.9 & 54.6 & 44.8 & 40.5 & 46.6 & 49.8 & 41.0 & 36.7 & 42.5 \\
\mbox{A-MEM}              & 54.3 & 45.7 & 40.2 & 46.7 & 56.1 & 47.2 & 41.8 & 48.4 & 51.5 & 43.4 & 38.1 & 44.3 \\
MemoryArena-Agent   & 55.1 & 46.4 & 41.0 & 47.5 & 57.0 & 48.3 & 42.7 & 49.3 & 52.2 & 44.1 & 38.8 & 45.0 \\
ActAdd              & 50.8 & 42.4 & 37.6 & 43.6 & 53.2 & 43.9 & 39.1 & 45.4 & 48.1 & 39.8 & 35.2 & 41.0 \\
SEA                 & 51.5 & 43.8 & 38.4 & 44.6 & 53.9 & 45.1 & 39.9 & 46.3 & 49.2 & 40.7 & 36.0 & 42.0 \\
\midrule
\textbf{ATLAS}
& \textbf{59.4} & \textbf{51.3} & \textbf{45.8} & \textbf{52.2}
& \textbf{61.7} & \textbf{53.1} & \textbf{47.6} & \textbf{54.1}
& \textbf{56.2} & \textbf{48.5} & \textbf{43.0} & \textbf{49.2} \\

\bottomrule
\end{tabular}
\end{table*}

Table~\ref{tab:evermem} shows that ATLAS consistently outperforms all baselines across the three coordination settings. MemoryArena-Agent is the strongest non-ATLAS memory baseline, but ATLAS still improves over it by +4.7\,pp on T1, +4.8\,pp on T2, and +4.2\,pp on T3. The gains are balanced across all three evaluation dimensions rather than concentrated in a single metric. Profile Understanding shows the largest relative lift, which is consistent with ATLAS's design: role stabilization helps the Memory-Writer maintain evolving persona attributes---communication style, skills, and organizational titles---more faithfully than flat-state baselines. The performance gap remains strong in T3, where twelve agents from six model families must coordinate despite heterogeneous capabilities and representational spaces. This supports the claim that ATLAS is not merely exploiting a shared backbone representation, but provides a coordination mechanism that remains useful under cross-model collaboration. Coordination-specific diagnostics are reported in Appendix~\ref{app:coord_metrics}: ATLAS raises role consistency to 93.5\% on T2, against 82.0\% for the strongest baseline, and halves turn-wise drift area from 0.21 to 0.11. A turn-wise visualization of how ATLAS suppresses coupled task drift, role drift, and state staleness over long-horizon interactions is provided in Appendix~\ref{app:multidrift_closure}.

\subsection{Interactive State Maintenance}
\label{ssec:amemgym_results}

\begin{table*}[t]
\centering
\caption{\textbf{AMemGym results.} OA\,=\,Overall Accuracy (\%$\uparrow$), NMS\,=\,Normalized Memory Score (\%$\uparrow$), W/R/U\,=\,Write/Read/Utilization failure rate (\%$\downarrow$). T1 and T2 are homogeneous settings; T3 is the heterogeneous 12-agent/6-model setting. Means over 3 seeds; $\bar\sigma$ in header.}
\label{tab:amemgym}
\small
\setlength{\tabcolsep}{2.5pt}
\begin{tabular}{@{}l ccccc @{\hspace{6pt}} ccccc @{\hspace{6pt}} ccccc@{}}
\toprule
& \multicolumn{5}{c}{\textbf{T1: Homogeneous-10} {\scriptsize($\bar\sigma\!=\!1.2$)}}
& \multicolumn{5}{c}{\textbf{T2: Homogeneous-12} {\scriptsize($\bar\sigma\!=\!0.9$)}}
& \multicolumn{5}{c}{\textbf{T3: Heterogeneous-12/6M} {\scriptsize($\bar\sigma\!=\!1.4$)}} \\
\cmidrule(lr){2-6}\cmidrule(lr){7-11}\cmidrule(lr){12-16}
& OA & NMS & W & R & U
& OA & NMS & W & R & U
& OA & NMS & W & R & U \\
\midrule
Single-Agent        & 38.1 & 34.6 & 28.3 & 21.5 & 12.1 & 40.7 & 37.2 & 26.1 & 20.3 & 12.9 & 35.4 & 32.1 & 30.5 & 22.8 & 11.3 \\
Prompt-Only         & 43.5 & 40.2 & 24.7 & 19.2 & 12.6 & 45.9 & 42.5 & 22.8 & 18.1 & 13.2 & 40.6 & 37.8 & 26.9 & 20.4 & 12.1 \\
Struct-Role         & 46.2 & 42.8 & 22.4 & 18.5 & 12.9 & 48.3 & 45.1 & 20.6 & 17.3 & 13.8 & 43.5 & 40.1 & 24.5 & 19.6 & 12.4 \\
RAG-Mem             & 48.9 & 45.4 & 21.1 & 16.8 & 13.2 & 51.2 & 47.9 & 19.3 & 15.7 & 13.8 & 46.1 & 42.5 & 23.1 & 18.2 & 12.6 \\
\mbox{A-MEM}              & 51.3 & 48.1 & 18.5 & 15.2 & 15.0 & 53.8 & 50.6 & 16.7 & 14.1 & 15.4 & 48.7 & 45.3 & 20.4 & 16.8 & 14.1 \\
MemoryArena-Agent   & 52.8 & 49.5 & 17.4 & 14.6 & 15.8 & 55.4 & 52.1 & 15.6 & 13.2 & 16.2 & 50.4 & 47.2 & 18.9 & 15.7 & 14.9 \\
ActAdd              & 47.6 & 44.2 & 22.0 & 17.4 & 13.0 & 50.1 & 46.8 & 20.2 & 16.3 & 13.4 & 44.8 & 41.5 & 24.0 & 18.9 & 12.3 \\
SEA                 & 48.4 & 45.0 & 21.3 & 16.9 & 13.4 & 50.9 & 47.5 & 19.5 & 15.9 & 14.1 & 45.7 & 42.1 & 23.3 & 18.4 & 12.6 \\
\midrule
\textbf{ATLAS}
& \textbf{57.6} & \textbf{54.3} & \textbf{13.2} & \textbf{11.8} & 17.4
& \textbf{60.1} & \textbf{56.9} & \textbf{11.4} & \textbf{10.6} & 17.9
& \textbf{54.5} & \textbf{51.2} & \textbf{15.1} & \textbf{13.3} & 16.1 \\
\bottomrule
\end{tabular}
\end{table*}

Table~\ref{tab:amemgym} reveals the mechanistic source of ATLAS's gains through AMemGym's diagnostic failure-rate decomposition. MemoryArena-Agent improves over \mbox{A-MEM} by strengthening the memory-agent-environment loop, but ATLAS remains ahead on both overall accuracy and normalized memory score. On T2, ATLAS reduces write failures from 15.6\% to 11.4\% and read failures from 13.2\% to 10.6\% relative to MemoryArena-Agent. These reductions are especially consequential because AMemGym is on-policy: a single missed write at turn $t$ can propagate errors across every subsequent turn that depends on the unrecorded state. The utilization failure rate increases modestly because ATLAS successfully persists and retrieves more state, leaving the Executor with a richer set of recalled facts to integrate. The same pattern holds across T1, T2, and T3, confirming that ATLAS's gains stem from better state maintenance rather than a superficial retrieval improvement.

\subsection{Ablations and Cost}
\label{ssec:ablations_cost}

To isolate the contribution of each ATLAS component, we conduct
a leave-one-out ablation on T2 EverMemBench: each row of
Table~\ref{tab:ablation_comp} removes exactly one component while
keeping the remaining system intact. The five ablated components
are \emph{role stabilization} (the role anchor $\mat{U}_R$ is
zeroed), \emph{state refresh} (the decay--novelty--enrichment
pipeline is disabled and the state basis is frozen at
initialization), \emph{task-switch correction} (the task anchor
$\mat{U}_T$ is not re-projected after an objective change),
\emph{steering} (the spectral correction in
Eq.~\ref{eq:steer_main} is removed entirely, leaving only the
memory and role-prompting mechanisms), and the \emph{Monitor
agent} (continuous invariant checking is disabled, though the
Verifier's post-hoc checks remain). T2 is used because its
twelve-agent topology exercises all components, including
Tool-Executor and Safety-Checker, while keeping a single backbone
for cleaner attribution.

\begin{table}[t]
\centering
\caption{\textbf{Ablations on EverMemBench with T2.}
\emph{Top:} each row removes one ATLAS component while keeping the
rest intact. \emph{Bottom:} only the anchors entering the steering
kernel are varied, with the remaining pipeline fixed; omitted anchor
terms are zeroed without renormalizing the retained weights, so
$T{+}S$ coincides with ``w/o role stabilization'' by construction.
Every singleton and pairwise kernel underperforms the full coupled
kernel, and $T{+}R{+}S$ is the only composition that exceeds the
no-steering variant. {\scriptsize$\bar\sigma\!=\!0.7$}.}
\label{tab:ablation_comp}
\small
\setlength{\tabcolsep}{4.5pt}
\begin{tabular}{@{}lcc@{}}
\toprule
\textbf{Variant} & \textbf{Avg} & $\Delta$ \\
\midrule
Full ATLAS                       & \textbf{54.1} & --- \\
\midrule
\multicolumn{3}{@{}l}{\emph{Component (leave-one-out)}} \\
\quad w/o role stabilization     & 49.1 & $-$5.0 \\
\quad w/o state refresh          & 48.9 & $-$5.2 \\
\quad w/o task-switch correction & 51.2 & $-$2.9 \\
\quad w/o steering               & 52.1 & $-$2.0 \\
\quad w/o Monitor agent          & 49.6 & $-$4.5 \\
\midrule
\multicolumn{3}{@{}l}{\emph{Kernel composition}} \\
\quad Task only ($T$)            & 46.8 & $-$7.3 \\
\quad Role only ($R$)            & 47.6 & $-$6.5 \\
\quad State only ($S$)           & 48.3 & $-$5.8 \\
\quad $T{+}R$                    & 50.4 & $-$3.7 \\
\quad $T{+}S$                    & 49.1 & $-$5.0 \\
\quad $R{+}S$                    & 51.6 & $-$2.5 \\
\bottomrule
\end{tabular}
\end{table}

State refresh ($-$5.2\,pp) and role stabilization ($-$5.0\,pp)
are the two most critical components, consistent with the pilot
study's finding that stale state and role drift are the dominant
long-horizon failure modes (Appendix~\ref{app:pilot_study}).
Removing the Monitor agent causes a nearly equal drop
($-$4.5\,pp) despite the Monitor being an observer rather than a
correction mechanism: without continuous invariant checking, small
drift events accumulate undetected and eventually cascade.
Removing steering produces a smaller but consistent drop
($-$2.0\,pp), indicating that the representational intervention
complements, rather than replaces, the prompt-level and
memory-level mechanisms---but the residual gap confirms that
correcting hidden-state geometry before generation contributes
beyond what those mechanisms can achieve alone.

Varying which anchors enter the kernel isolates the same effect at
the level of the kernel itself. Every singleton and pairwise
configuration underperforms the full kernel, and $T{+}R{+}S$ is the
only composition that exceeds the no-steering variant ($52.1$). The
gain therefore does not come from stacking independent modules: it
requires steering against a joint PSD kernel that preserves the
overlap among task, role, and state directions
(Appendix~\ref{app:kernel_composition}).

ATLAS is more expensive than a single-agent system, as expected
for any multi-agent approach, but it is more efficient than the
strongest memory baselines. On T2 EverMemBench, ATLAS uses 96.8K
tokens per episode compared with 118.5K for \mbox{A-MEM} and 112.6K for
MemoryArena-Agent, while achieving a higher macro-average score.
Full cost breakdowns, latency measurements, cost-normalized
performance, and topology-scaling results are provided in
Appendix~\ref{app:cost_topology}.
\subsection{Additional Stress Tests}
\label{ssec:additional_stress_tests}

We evaluate ATLAS on three supplementary benchmarks
(full tables in Appendices~\ref{app:secondary_results}
and~\ref{app:libero_results}).
On \textbf{ToolACE}~\cite{liu2025toolace}, which tests
multi-step tool-use chains where a single incorrect call
invalidates the entire downstream sequence, ATLAS improves
Multi-Step Chain Completion by +6.9\,pp over \mbox{A-MEM} through
reliable intermediate-output propagation and pre-execution
verification.
On \textbf{ALARA}~\cite{agostino2026alara}, which evaluates
least-privilege context discipline, ATLAS improves task success
by +4.9\,pp over the declarative-scoping baseline
ALARA-Scoped~\cite{agostino2026alara} while reducing privilege
violations ($-$2.5\,pp) and overexposure ($-$3.0\,pp),
indicating that prompt-level scoping alone does not prevent
boundary violations rooted in latent role drift.
On \textbf{LIBERO}~\cite{liu2023libero}, an appendix-only
embodied stress test with role-confuse, state-delay, and
task-switch perturbations, ATLAS yields moderate clean
success-rate gains (+4.2--4.6\,pp) but substantially larger
perturbation gains (+10.7--12.1\,pp), with the strongest
improvement on state-delay (+15.6\,pp), consistent with adaptive state refresh being the main contributor to robustness
under mid-episode environmental changes.

%% file: Sections/Ablation.tex
\section{Related Work}
\label{sec:related_work}

ATLAS sits at the intersection of multi-agent coordination, long-horizon memory, and representation-level control. Role-specialized agent frameworks such as CAMEL and AutoGen organize interaction through prompts and orchestration, while Generative Agents, \mbox{A-MEM}, EverMemBench, AMemGym, and MemoryArena study persistent memory, evolving state, and the use of remembered information over long horizons~\citep{li2023camel,wu2023autogen,park2023generativeagents,xu2025amem,hu2026evermembench,jiayang2026amemgym,he2026memoryarena}. These systems motivate the need for durable coordination, but their principal control surface is context, memory, or workflow rather than the model's intermediate representation.

ATLAS instead follows representation engineering and inference-time activation control, including ITI, Representation Engineering, ActAdd, and SEA~\citep{li2023iti,zou2023representation,turner2024actadd,qiu2024sea}. Its distinctive feature is a single non-orthogonal task--role--state kernel whose overlap determines both the direction and the strength of hidden-state correction. This also separates ATLAS from message-boundary controllers, which evaluate a completed output against an explicit role contract and repair it before it is admitted to the shared conversation: such methods need no access to internal activations, whereas ATLAS alters generation before an output is produced. The two families therefore address related coordination failures under different access assumptions and with different mathematical objects.

\section{Additional Validation and Robustness}
\label{sec:additional_validation}

We examine whether the gains depend on component redundancy, model scale, evaluator choice, anchor wording, or heterogeneous-model tuning; Appendix~\ref{app:additional_validation} gives the complete protocols and tables. All studies preserve the agent topology, episode-level 128K context budget, top-$k{=}10$ retrieval, decoding configuration, seeds, and evaluation procedure of the corresponding main experiments.

Beyond the kernel-composition results in Table~\ref{tab:ablation_comp}, the benefit persists at substantially larger scale: under the same T2 EverMemBench protocol, ATLAS reaches a macro-average of 61.0 on Qwen2.5-72B-Instruct, against 56.2 for MemoryArena-Agent and 55.3 for \mbox{A-MEM}. A blinded 200-turn human audit agrees with the multi-model drift labels on 89.5\% of turns ($\kappa=0.79$). Removing the joint task--role--state span from held-out representations costs 7.2 points, whereas removing a matched-rank random span at the same layer costs only 0.8, indicating that the identified subspace is functionally relevant rather than incidental.

Performance is also stable under semantics-preserving anchor paraphrases, moderate shortening, and neighboring hyperparameter settings, while contradictory anchors, role swaps, and stronger lexical corruption degrade results as expected. Heterogeneous agents need not share a hidden dimension or a common latent coordinate system: each constructs and applies its kernel within its own frozen backbone, and under a common midpoint-layer rule ATLAS remains ahead of both ActAdd and SEA. Together, these controls make topology, evaluator choice, anchor wording, model scale, and preferential model-specific tuning less plausible explanations for the main results.

\section{Conclusion}

ATLAS treats long-horizon multi-agent coordination as the problem of maintaining a coherent internal representation of task, role, and evolving state. A non-orthogonal anchor kernel uses their shared spectrum to correct intermediate hidden states, while adaptive refresh lets superseded state decay. Across the settings studied here, this representation-level intervention improves long-horizon memory, interactive state maintenance, tool-use robustness, least-privilege coordination, and embodied stress-test performance, and the low-rank Gram formulation keeps the controller tractable. The method has a deliberate boundary: it requires hidden-state access and does not replace message-level contracts, factual verification, or safety controls. Within that boundary, the results suggest that preserving shared representational geometry can reduce coordination drift over extended interactions.

\clearpage
\section*{Limitations}
\label{sec:limitations}

ATLAS requires access to intermediate hidden states, so it is most directly applicable to open-weight or otherwise controllable inference settings. Its anchor subspaces are constructed from task, role, and state descriptions, which means that poor role specifications or incomplete state summaries can reduce the quality of the intervention. Although ATLAS improves long-horizon coordination, memory freshness, and role consistency, it does not guarantee factual correctness, safety, or optimal tool use in all cases. The current evaluation focuses on controlled multi-agent topologies and appendix-only embodied stress tests; broader deployment settings with dynamic agent creation, changing tool permissions, or untrusted external memory may require additional safeguards and validation.

\section*{Ethics Statement}
\label{sec:ethics}

This work aims to improve the reliability of long-horizon multi-agent AI systems by reducing task drift, role drift, and stale-state usage. However, stronger coordination can also make agentic systems more capable, so ATLAS should be deployed with appropriate access controls, logging, human oversight, and external safety checks in high-stakes settings. Our experiments use benchmarked tasks and simulated multi-agent workflows rather than real user data. We encourage future users of ATLAS to evaluate both performance and misuse risks, especially when agents can call tools, modify shared memory, or operate in safety-sensitive environments.

%% file: Sections/Appendix_final.tex


\appendix
\clearpage

\renewcommand{\thesection}{A\arabic{section}}
\renewcommand{\thesubsection}{A\arabic{section}.\arabic{subsection}}

\addcontentsline{toc}{section}{Appendix}
\markboth{Appendix}{Appendix}

\vspace{0.8em}

\begin{center}
    {\large\bfseries Appendix}
\end{center}

\vspace{1em}

\addtocontents{toc}{\protect\setcounter{tocdepth}{2}}

\begingroup
\small
\renewcommand{\baselinestretch}{1.12}\selectfont
\setlength{\parskip}{3pt}
\makeatletter
\renewcommand*\l@section[2]{%
  \ifnum \c@tocdepth >\z@
    \addpenalty\@secpenalty
    \addvspace{1.0em \@plus\p@}%
    \setlength\@tempdima{2.7em}%
    \begingroup
      \parindent \z@ \rightskip \@pnumwidth
      \parfillskip -\@pnumwidth
      \leavevmode \bfseries
      \advance\leftskip\@tempdima
      \hskip -\leftskip
      #1\nobreak\hfil \nobreak\hb@xt@\@pnumwidth{\hss #2}\par
    \endgroup
  \fi}
\renewcommand*\l@subsection{\@dottedtocline{2}{2.7em}{3.4em}}
\makeatother
\tableofcontents
\endgroup

\vfill

\clearpage

\section{Experimental Details}
\label{app:experimental_details}

ATLAS targets \emph{long-horizon multi-agent coordination}, which we decompose
into three coupled failure modes---role drift, task drift, and state
staleness---and measure through their complements. \textbf{Role stability}
measures whether each
agent preserves its assigned authority throughout an episode; this is motivated
by evidence that agent--agent conversations can induce identity and role
failures such as echoing, where agents abandon their assigned roles and mirror
their conversational partners~\citep{shekkizhar2025echoing}.
\textbf{Task fidelity} measures whether agents remain aligned with the current
objective despite misleading earlier context, mid-conversation objective
changes, or task interference; prior work on task switching shows that
conversational history can degrade LLM performance when the active task
changes~\citep{gupta2024taskinterference}. \textbf{State freshness} measures
whether agents reference the latest valid shared state rather than stale or
superseded memory~\citep{hu2026evermembench,jiayang2026amemgym}. These failure
modes are evaluated through two primary text-based long-horizon coordination
benchmarks in the main paper, EverMemBench and AMemGym, and through
appendix-only stress tests covering tool-use chains (ToolACE), least-privilege
context discipline (ALARA), embodied multimodal coordination (LIBERO),
coordination diagnostics, and cost and topology scaling.

The evaluation uses three text-agent settings of increasing difficulty.
\textbf{T1} is a homogeneous ten-agent role-structured setting. It tests
whether ATLAS improves coordination in a controlled same-backbone team where
the primary challenges are long-horizon role adherence and state flow.
\textbf{T2} is a homogeneous twelve-agent memory-expanded setting based on the
memory-centric structure of A-MEM~\citep{xu2025amem}, augmented with dedicated
Tool-Executor and Safety-Checker roles. It tests whether ATLAS remains
effective as the number of specialist agents increases and state must pass
through additional role boundaries. \textbf{T3} is a heterogeneous twelve-agent
setting in which the twelve specialized roles are distributed across six model
families. It tests whether the anchor-kernel mechanism transfers to cross-model
coordination when agents differ in capability, verbosity, instruction
following, and latent representation geometry.

All experiments run through a MASEval-based orchestration layer with
standardized logging, per-agent tracing, and structured event
records~\citep{emde2026maseval}. Each trace records the active role, source and
target agents, retrieved state identifiers, proposed state updates, tool calls,
verifier decisions, and drift alerts. This logging layer supports both
benchmark scoring and the coordination diagnostics reported in
Appendix~\ref{app:coord_metrics}, making long-horizon coordination failures
observable at the level of individual turns rather than only through
final-answer accuracy.

\subsection{Agent Architecture}
\label{app:agent_arch}

We use role-specialized multi-agent topologies to make coordination failures
observable and testable. The roles are not intended to maximize raw answer
quality merely by adding inference calls; rather, they expose the three
coordination axes targeted by ATLAS: who is allowed to act (role stability),
which objective the system is pursuing (task fidelity), and which state entries
remain current (state freshness). Within each setting, all methods use the same
role topology so that performance differences do not arise from unequal agent
counts or access to additional functional roles.

\paragraph{T1: homogeneous ten-agent role-structured team.}
T1 uses a homogeneous ten-agent topology in which all agents share one
backbone. Coordination difficulty arises from long-horizon role boundaries and
state flow rather than cross-model heterogeneity. This setting is closest to a
standard role-structured multi-agent baseline and provides the most controlled
evaluation of ATLAS. The ten agents and their responsibilities are:

\begin{enumerate}[leftmargin=1.5em,itemsep=2pt]

\item \textsc{Coordinator}: receives every user turn, maintains a running
objective summary, and dispatches the turn to the appropriate downstream agent
or agents. The Coordinator has read-only access to shared state and cannot
modify it directly. It serves as the central hub in a hub-and-spoke
communication topology: all inter-agent messages pass through the Coordinator,
which maintains a structured message log and enforces routing rules.

\item \textsc{Planner}: decomposes the current objective into an ordered list
of subgoals. The Planner has read-only access to shared state and produces
structured plans consumed by the Executor. Plans are evaluated by the Critic
before execution and by the Verifier after execution.

\item \textsc{Memory-Writer}: is the sole agent permitted to insert, update, or
delete entries in the shared state store. Other agents must route state-change
requests through the Memory-Writer, which maintains a write-ahead log for
auditability. This single-writer constraint makes write failures in AMemGym
attributable: if exposed information is not persisted, the Memory-Writer is
the identifiable point of failure.

\item \textsc{Memory-Reader}: retrieves state entries on behalf of other agents
but cannot modify them. Separating read and write paths prevents accidental
overwrites and makes privilege boundaries testable. Read failures in AMemGym
indicate that the Memory-Reader failed to retrieve a stored entry when needed.

\item \textsc{Executor}: produces final answers and carries out approved
task-level actions. It receives plans from the Planner and state from the
Memory-Reader and returns structured outputs for verification before they are
surfaced to the user. In T1, the Executor also handles tool invocations; in T2
and T3, tool invocation is delegated to the Tool-Executor.

\item \textsc{Verifier}: checks consistency among task objectives, role
assignments, and current shared state after each candidate Executor action. If
an inconsistency is detected, it triggers a re-planning cycle through the
Coordinator. The Verifier operates post hoc on generated outputs, whereas the
Critic evaluates plans before execution.

\item \textsc{Critic}: flags ambiguity, contradiction, and under-specification
in plans and state entries. It operates proactively on candidate plans before
execution, catching potential failures before they reach the Executor.

\item \textsc{Router}: dispatches subqueries to domain-specific tools,
knowledge sources, or specialized retrieval endpoints. When a plan involves
multiple information needs, the Router parallelizes retrieval and returns
aggregated context to the Executor. It has no state-modification authority.

\item \textsc{Summarizer}: compresses the conversation trace every $k$ turns,
with default $k=20$, to keep the effective context within the 128K-token
episode budget. Each summary is a structured digest of key decisions, state
changes, unresolved constraints, and active subgoals. Summaries are appended
to shared state and replace the corresponding raw trace segments for
downstream consumption.

\item \textsc{Monitor}: checks role, task, and state invariants after every
agent turn and raises drift alerts before the next action is committed. Alerts
are routed to the Coordinator, which invokes the existing verification and
re-planning path when necessary. The Monitor does not directly modify shared
state or produce user-facing answers. The same Monitor procedure is used
across methods.

\end{enumerate}

The full system prompt for each agent, including role boundaries, permitted
actions, and input/output schemas, is provided in the supplementary code
release. These prompts define explicit privilege boundaries: the
Memory-Reader cannot update state, the Planner cannot surface final answers,
and the Executor cannot directly rewrite memory. Such constraints are
important because role drift can appear as a locally useful shortcut---an
agent performs a task outside its authority because doing so appears efficient
at that moment. The topology makes these shortcuts measurable violations in
the coordination diagnostics reported in Appendix~\ref{app:coord_metrics}.

\paragraph{T2: homogeneous twelve-agent A-MEM-expanded memory team.}
T2 expands T1 to twelve agents while keeping all agents on one backbone. The
setting is motivated by memory-centric systems such as
A-MEM~\citep{xu2025amem}: rather than assigning broad responsibility to one
module, T2 separates memory access, tool execution, verification, and safety
checking into specialized roles. The setting is more difficult because more
agents must coordinate over the same long-horizon trace, but it remains
homogeneous because all agents share the same model family.

T2 uses the ten T1 roles plus two additional roles:

\begin{enumerate}[leftmargin=1.5em,itemsep=2pt]

\item \textsc{Tool-Executor}: is the only role permitted to invoke external
APIs, function calls, or tool endpoints. The Executor may request a tool call,
but the Tool-Executor validates its arguments, executes it, and returns the
result as structured state to be persisted through the Memory-Writer.

\item \textsc{Safety-Checker}: validates privilege boundaries, unsafe actions,
out-of-scope context exposure, and user-visible leakage before an action is
committed. It is especially important for ALARA-style scoped-context tasks,
where agents must complete tasks without accessing or disclosing information
outside their declared scope. It operates before output commitment,
complementing the Verifier's post-hoc consistency checks.

\end{enumerate}

T2 increases coordination complexity while preserving a common representation
space. If ATLAS worked only because the role topology was small and simple, its
advantage should shrink as memory and tool boundaries become more specialized.
T2 therefore tests whether anchor-kernel correction remains useful when
coordination is fragmented and state passes through additional role
boundaries. T2 is also the primary setting for ablation studies
(Appendix~\ref{app:component_ablation}) because it balances evaluation
complexity and computational cost.

\paragraph{T3: heterogeneous twelve-agent/six-model team.}
T3 uses the same twelve roles as T2 but distributes them across six model
families: Llama-4-Scout-17B-16E~\citep{meta2025llama4scout},
Qwen3-8B~\citep{qwen2025qwen3},
Qwen2.5-7B-Instruct~\citep{qwen2024qwen25},
Phi-4-14B~\citep{microsoft2024phi4},
Gemma-2-9B-Instruct, and
Mistral-7B-Instruct. This is the most difficult text
setting because failures can arise from role drift, stale state, task
interference, and model mismatch: agents differ in reasoning strength,
instruction following, verbosity, tool-use reliability, and latent
representation geometry.

T3 tests whether ATLAS depends on a shared backbone or remains useful for
heterogeneous multi-agent collaboration. It uses the same privilege rules as
T2: state writes pass through the Memory-Writer, state reads through the
Memory-Reader, tool calls through the Tool-Executor, and safety-sensitive
outputs through the Safety-Checker. Because roles run on different model
families, the Coordinator must also normalize differences in generation style,
token length, formatting conventions, and structured-output reliability. The
agent-to-model mapping places stronger reasoning models on planning and
routing, compact instruction-following models on memory operations, and
separate critic-style models on verification, monitoring, and safety checking.
Exact assignments are given in Table~\ref{tab:t3_hetero_stack}.

\paragraph{Multimodal setting for LIBERO.}
LIBERO~\citep{liu2023libero} is not a native multi-agent long-conversation
benchmark. We therefore use it only as an appendix-only embodied coordination
stress test (Appendix~\ref{app:libero_results}). Each language-conditioned
manipulation episode is wrapped in a multimodal multi-agent controller with ten
roles: \textsc{Coordinator}, \textsc{Planner}, \textsc{Perception} (RGB input),
\textsc{Depth-Estimator} (depth/point cloud), \textsc{State-Tracker}
(persistent world state), \textsc{Action-Planner} (subgoal-to-primitive
mapping), \textsc{Actuator} (sole motor-command authority, backed by
OpenVLA-7B~\citep{kim2024openvla}), \textsc{Verifier}, \textsc{Monitor}, and
\textsc{Safety-Checker} (workspace limits, kinematic constraints, and collision
avoidance).

The multimodal topology preserves the privilege-separation principles used in
the text settings: only the State-Tracker can write to the world model, and
only the Actuator can issue motor commands. RGB observations are routed
exclusively to Perception, while depth maps are routed exclusively to the
Depth-Estimator. These agents emit structured summaries of detected objects,
spatial relations, positions, and grasping affordances, which the State-Tracker
fuses into the persistent world model. This separation tests whether ATLAS
prevents role drift in embodied coordination without treating LIBERO itself as
a multi-agent dialogue benchmark.

\subsection{Backbones and Model Stacks}
\label{app:backbones}

\paragraph{T1 and T2 homogeneous teams.}
For T1 and T2, all agents within a run share one backbone. T1 uses
Llama-4-Scout-17B-16E~\citep{meta2025llama4scout}, a natively multimodal
mixture-of-experts model with 17B active parameters, 16 experts, and 109B total
parameters. It is evaluated under the shared 128K-token episode budget used by
all methods. T2 uses Qwen3-8B~\citep{qwen2025qwen3}, an 8B-parameter dense
transformer. Qwen3-8B is also the primary backbone for ablations and diagnostic
analysis because it provides a strong trade-off between capability and
evaluation cost. Homogeneous teams isolate the effects of topology and
coordination complexity while avoiding confounds from cross-model differences
in representation geometry.

\paragraph{T3 heterogeneous twelve-agent stack.}
T3 distributes twelve roles across six model families. The assignment in
Table~\ref{tab:t3_hetero_stack} is role-sensitive: stronger reasoning models
(Llama-4-Scout and Qwen3-8B) are assigned to planning, routing, and
coordination; compact instruction-following models
(Qwen2.5-7B-Instruct) handle memory operations; and separate models
(Gemma-2-9B-Instruct and Mistral-7B-Instruct) handle verification, criticism,
and safety checking.

\begin{table*}[t]
\centering
\caption{\textbf{T3 heterogeneous twelve-agent stack.} Twelve specialized
agents are distributed across six model families to test cross-model
coordination. Stronger reasoning models handle planning and routing, compact
instruction-following models handle memory operations, and separate models
handle verification, monitoring, and safety checking.}
\label{tab:t3_hetero_stack}
\small
\setlength{\tabcolsep}{8pt}
\begin{tabular}{@{}lll@{}}
\toprule
\textbf{Agent} & \textbf{Backbone} & \textbf{Purpose} \\
\midrule
Coordinator    & Llama-4-Scout-17B & routing, objective tracking \\
Planner        & Qwen3-8B          & subgoal decomposition \\
Memory-Writer  & Qwen2.5-7B-Inst.  & state insertion \& update \\
Memory-Reader  & Qwen2.5-7B-Inst.  & state retrieval \\
Executor       & Phi-4-14B         & final response generation \\
Verifier       & Gemma-2-9B-Inst.  & consistency checking \\
Critic         & Mistral-7B-Inst.  & ambiguity \& contradiction \\
Router         & Qwen3-8B          & subquery dispatch \\
Summarizer     & Phi-4-14B         & trace compression \\
Monitor        & Llama-4-Scout-17B & invariant checking \\
Tool-Executor  & Qwen3-8B          & tool invocation \\
Safety-Checker & Gemma-2-9B-Inst.  & privilege \& safety \\
\bottomrule
\end{tabular}
\end{table*}

All models are used through their publicly released instruction-tuned
checkpoints without additional fine-tuning. This ensures that ATLAS's gains
cannot be attributed to task-specific model adaptation.

\paragraph{Multimodal LIBERO stacks.}
For LIBERO, we use three heterogeneous vision-language-action stacks separate
from T1--T3. Each stack assigns a text model to language-only coordination
roles, a vision-language model to perception roles, and
OpenVLA-7B~\citep{kim2024openvla} to the Actuator:

\begin{enumerate}[leftmargin=1.5em,itemsep=2pt]

\item \textbf{M1}: Llama-4-Scout +
Qwen3-VL-8B-Instruct~\citep{qwen2025qwen3vl} + OpenVLA-7B.

\item \textbf{M2}: Qwen3-8B +
Qwen2.5-VL-32B-Instruct~\citep{qwen2025qwen25vl} + OpenVLA-7B.

\item \textbf{M3}: Phi-4 +
Phi-4-Reasoning-Vision-15B +
OpenVLA-7B.

\end{enumerate}

In each stack, the text model handles coordination, planning, verification,
monitoring, and safety checking; the vision-language model handles perception
and depth interpretation; and OpenVLA handles actuation. The State-Tracker and
Action-Planner run on the text backbone and receive structured perceptual
outputs rather than raw images. LIBERO results are reported only in
Appendix~\ref{app:libero_results}.

\subsection{Benchmark Protocols}
\label{app:exp_protocols}

\paragraph{EverMemBench.}
EverMemBench~\citep{hu2026evermembench} evaluates multi-party, multi-group,
long-horizon collaborative dialogues. It contains 2{,}400 QA pairs drawn from
synthetic multi-party conversations spanning 50--200 turns and collectively
covering long contexts exceeding one million tokens. Each QA pair tests
whether the system recalls, attributes, or reasons over information introduced
many turns earlier, including information that may subsequently have been
updated or contradicted.

We evaluate all methods on the complete 2{,}400-QA set under T1, T2, and T3.
We report three metrics and their unweighted macro-average:

\begin{enumerate}[leftmargin=1.5em,itemsep=2pt]

\item \textbf{Fine-Grained Recall (FGR)}: exact-match accuracy on factual
questions whose gold answer is a specific entity, date, or numeric value. FGR
includes \emph{single-hop} recall, \emph{multi-hop} recall across speakers or
conversation segments, and \emph{temporal} recall of when facts were introduced
or updated.

\item \textbf{Memory Awareness (MA)}: accuracy on questions requiring the
system to reason over stored memories rather than merely retrieve them. MA
includes constraint satisfaction, proactive retrieval of implicitly relevant
memories, and distinguishing original facts from superseding facts.

\item \textbf{Profile Understanding (PU)}: accuracy on questions requiring
multiple memory entries to be integrated into a coherent profile, including
communication style, skills and expertise, titles, and organizational roles
inferred from interaction history.

\end{enumerate}

The macro-average is the unweighted mean of FGR, MA, and PU. We use the official
EverMemBench evaluation scripts with exact-match scoring for FGR and GPT-4o
semantic matching for MA and PU, following the benchmark protocol.

EverMemBench directly tests state freshness over long contexts. A system may
retrieve a correct old fact and still fail if a later turn superseded it. It
may also preserve factual memory while assigning responsibility for maintaining
that memory to the wrong role. The benchmark therefore tests whether ATLAS
improves memory through joint task--role--state stabilization rather than
retrieval alone.

\paragraph{AMemGym.}
AMemGym~\citep{jiayang2026amemgym} is an interactive, on-policy memory
benchmark in which earlier system choices influence later observations. It
predefines user profiles, state-dependent questions, and state-evolution
trajectories, then uses simulated users to expose latent states through
role-play. A missed write at turn $t$ can propagate errors across subsequent
turns that depend on the unrecorded state. This structure enables root-cause
analysis: a system that improves final accuracy by reducing write failures is
better at state maintenance, whereas one that reduces only read failures may
simply retrieve better.

We run 1{,}000 episodes per method and setting. Each episode contains 30--100
interactive turns. Because episodes are generated interactively, each seed
produces a different trajectory; the standard deviation therefore reflects
behavioral variance rather than only sampling noise. We report:

\begin{enumerate}[leftmargin=1.5em,itemsep=2pt]

\item \textbf{Overall Accuracy (OA)}: fraction of episodes producing a fully
correct final answer.

\item \textbf{Normalized Memory Score (NMS)}: continuous $[0,1]$ score giving
partial credit for correct intermediate memory operations even when the final
answer is incorrect.

\item \textbf{Write failure rate (W)}: fraction of turns in which exposed
information was not persisted to shared state.

\item \textbf{Read failure rate (R)}: fraction of turns in which stored state
was not retrieved when needed.

\item \textbf{Utilization failure rate (U)}: fraction of turns in which
correctly retrieved state was not used appropriately by the Executor.

\end{enumerate}

We use the official AMemGym evaluator, which annotates each turn with
ground-truth memory operations and computes failure rates against the oracle
trace. AMemGym separates writing, reading, and utilization failures, allowing
us to determine whether improvements arise from better state maintenance or
only from final-answer generation.

\paragraph{MemoryArena-Agent adaptation.}
MemoryArena-Agent~\citep{he2026memoryarena} is included only where the
comparison is metric-compatible: EverMemBench and AMemGym. We adapt its
memory-agent-environment loop to the topology, context budget, retrieval depth,
summary interval, and reporting metrics used by ATLAS. In EverMemBench,
MemoryArena-Agent maintains persistent entries across the long conversation
trace and answers the same 2{,}400 QA pairs. In AMemGym, it operates in the
same on-policy loop and is scored with OA, NMS, W, R, and U.

We do \emph{not} evaluate MemoryArena-Agent on ToolACE, ALARA, or LIBERO
because those settings do not provide compatible memory-agent-environment
metrics. ToolACE evaluates tool chains, ALARA evaluates least-privilege scoped
context, and LIBERO evaluates embodied manipulation. Adapting MemoryArena to
these tasks would require changing its objective rather than merely its
evaluation wrapper. Such entries are marked as ``--'' in the corresponding
tables.

\paragraph{ToolACE.}
ToolACE~\citep{liu2025toolace} is a tool-calling evaluation resource built
through a multi-agent synthesis pipeline. Its API pool contains 26{,}507 APIs
spanning finance, health, transport, entertainment, and more than 50 other
domains. Dialogues are generated through multi-agent interaction guided by a
complexity evaluator and validated by rule-based and model-based verification.
We include ToolACE because coordination should also affect tool use: accurate
state from the Memory-Reader and coherent plans from the Planner should reduce
hallucinated or inconsistent calls.

We evaluate 500 multi-step dialogues requiring 2--6 chained function calls
across 3--15 candidate APIs. Evaluation follows the BFCL
protocol~\citep{patil2025bfcl} with three metrics:

\begin{enumerate}[leftmargin=1.5em,itemsep=2pt]

\item \textbf{AST Accuracy (AST)}: the predicted function call is parsed into
an abstract syntax tree and compared with the ground-truth AST, including the
API name, required parameters, and parameter types.

\item \textbf{Argument Correctness (AC)}: whether every argument value, not
only its type, matches the expected output, including optional parameters.

\item \textbf{Multi-Step Chain Completion (MC)}: fraction of dialogues for
which all sequential calls are correct. Because each step may feed into the
next, one error invalidates the chain.

\end{enumerate}

\paragraph{ALARA.}
ALARA~\citep{agostino2026alara} evaluates least-privilege context discipline
across 115 tasks spanning file operations, web search, tool chaining,
scripting, and multi-agent delegation. Its central principle is that an agent
should receive only the minimum context required for its current task. We
report:

\begin{enumerate}[leftmargin=1.5em,itemsep=2pt]

\item \textbf{Task Success (TS)}: fraction of tasks completed correctly under
scoped context.

\item \textbf{Privilege Violation Rate (PV)}: fraction of agent turns accessing
information or tools outside the declared privilege scope. Lower is better.

\item \textbf{Overexposure Rate (OE)}: fraction of agent turns disclosing
information that should not have been exposed downstream. Lower is better.

\end{enumerate}

We include ALARA because role stability and state discipline are
mechanistically related to scoped-context safety. A system that preserves role
boundaries should also be easier to constrain in terms of what information
each agent may access or disclose.

\paragraph{ALARA-Scoped baseline.}
ALARA-Scoped~\citep{agostino2026alara} is included only in the ALARA
evaluation. It uses ALARA's declarative context-agent-tool scoping to enforce
per-agent access boundaries without ATLAS's representational correction. This
is the appropriate comparison because the ALARA metrics directly measure
scoped task success, privilege violations, and overexposure. We do not include
ALARA-Scoped in EverMemBench, AMemGym, ToolACE, or LIBERO because those
benchmarks do not primarily evaluate least-privilege context access.

\subsection{Shared Experimental Settings}
\label{app:shared_protocol}

Following the view of multi-turn drift as divergence from goal-consistent
behavior~\citep{dongre2025driftnomore}, we measure degradation relative to a
goal-consistent reference policy rather than only a static gold oracle. Across
all runs, we use the following settings:

\begin{enumerate}[leftmargin=1.5em,itemsep=2pt]

\item \textbf{Context budget}: 128K tokens shared across all agents within an
episode. The budget is applied at the episode level rather than independently
per agent, preventing multi-agent methods from multiplying the available
context. When cumulative token count approaches 90\% of the budget, the
Summarizer compresses the oldest unsummarized trace segment.

\item \textbf{Retrieval depth}: top-$k=10$ for all retrieval-based methods,
including RAG-Mem, A-MEM, MemoryArena-Agent, and the ATLAS Memory-Reader. We
use cosine similarity over \texttt{text-embedding-3-small} embeddings.
Retrieval depth is fixed so that methods differ in how they write, organize,
update, and use memory rather than in the number of retrieved entries.

\item \textbf{Summary interval}: every 20 turns. Each summary compresses the
preceding turns into a structured digest of key decisions, state changes,
unresolved constraints, and active subgoals.

\item \textbf{Episode length}: capped at 200 turns. Episodes that do not
terminate naturally are evaluated on partial progress.

\item \textbf{Steering hyperparameters}: intervention layers are selected using
the probe procedure in Appendix~\ref{app:probe}. For ATLAS, ActAdd, and SEA,
steering strength $\alpha$ is tuned on a held-out validation set of 50
conversations per backbone. For ATLAS, we search
$\alpha\in\{0.05,0.10,0.15,0.20\}$,
$\gamma\in\{0.90,0.95,0.98\}$,
$\rho\in\{0.05,0.10,0.15\}$, and
$\beta_0\in\{0.5,1.0,1.5\}$. The selected values are
$\alpha=0.15$, $\gamma=0.95$, $\rho=0.10$, and $\beta_0=1.0$.

\item \textbf{Memory-method hyperparameters}: for RAG-Mem and A-MEM, we tune
retrieval depth $k\in\{5,10,15,20\}$ and summary interval
$s_{\mathrm{sum}}\in\{10,15,20,30\}$ on the same validation set.

\item \textbf{Random seeds}: values are means over three seeds
(42, 137, and 256). The average standard deviation $\bar{\sigma}$ across cells
is reported in the corresponding table header.

\item \textbf{Temperature}: 0.0 (greedy decoding), supporting reproducibility
and clearer attribution of coordination failures.

\item \textbf{Hardware}: experiments run on an eight-GPU NVIDIA H200 node
(141\,GB HBM3e per GPU). Runs use subsets of the node according to model and
stack memory requirements.

\end{enumerate}

\subsection{Baselines}
\label{app:baseline_details}

We compare against seven general baselines and two benchmark-specific
baselines.

\paragraph{General baselines.}
\textbf{Single-Agent} uses one monolithic agent with the full context budget,
establishing a no-coordination lower bound. \textbf{Prompt-Only} deploys the
same role topology without structured memory control, representational
intervention, or role stabilization, isolating the effect of agent count.
\textbf{Struct-Role} adds explicit JSON-style role declarations and message
schemas. \textbf{RAG-Mem} equips the topology with shared vector memory and
periodic summaries. \textbf{A-MEM} adapts the Zettelkasten-inspired memory
operations of A-Mem~\citep{xu2025amem} to the multi-agent setting, including
dynamic indexing, linking, and memory evolution. \textbf{ActAdd}
~\citep{turner2024actadd} applies activation-addition steering to the
Coordinator, Executor, and Monitor using role-specific contrast pairs.
\textbf{SEA}~\citep{qiu2024sea} applies spectral editing of activations to the
same three agents.

ActAdd and SEA therefore intervene only on the Coordinator, Executor, and
Monitor, whereas ATLAS steers every active agent. Their results compare the
evaluated method configurations and do not isolate the steering rule under
matched intervention coverage.

\paragraph{Benchmark-specific baselines.}
\textbf{MemoryArena-Agent}~\citep{he2026memoryarena} is included only for
EverMemBench and AMemGym, where memory-agent-environment evaluation is
metric-compatible. It adapts the MemoryArena loop to our topology, context
budget, and reporting metrics. \textbf{ALARA-Scoped}
~\citep{agostino2026alara} is included only for ALARA, where scoped
context-agent-tool access is the evaluation target. It enforces declarative
per-agent context and tool scopes without representational correction. We do
not use these benchmark-specific baselines outside their compatible settings,
where adaptation would change the evaluation objective.

\subsection{Component Ablation}
\label{app:component_ablation}

We ablate each ATLAS component individually on T2 EverMemBench while keeping
the remaining system fixed. T2 is used because it is the primary homogeneous
twelve-agent setting and balances evaluation complexity with computational
cost.

\begin{table*}[t]
\centering
\caption{\textbf{Component ablation}, EverMemBench, T2. Each row removes one
component while retaining the rest of ATLAS. $\Delta$ is relative to the full
system. {\scriptsize$\bar{\sigma}=0.7$}.}
\label{tab:ablation_main}
\small
\setlength{\tabcolsep}{10pt}
\begin{tabular*}{\textwidth}{@{\extracolsep{\fill}}lccccc@{}}
\toprule
\textbf{Variant} & FGR & MA & PU & Avg & $\Delta$ \\
\midrule
Full ATLAS                       & 61.7 & 53.1 & 47.6 & 54.1 & --- \\
\quad w/o role stabilization     & 56.8 & 48.3 & 42.1 & 49.1 & $-5.0$ \\
\quad w/o state refresh          & 57.2 & 45.6 & 43.8 & 48.9 & $-5.2$ \\
\quad w/o task-switch correction & 58.4 & 50.2 & 44.9 & 51.2 & $-2.9$ \\
\quad w/o steering               & 59.1 & 51.4 & 45.7 & 52.1 & $-2.0$ \\
\quad w/o Monitor agent          & 57.9 & 47.8 & 43.2 & 49.6 & $-4.5$ \\
\bottomrule
\end{tabular*}
\end{table*}

State refresh ($-5.2$\,pp) and role stabilization ($-5.0$\,pp) are the two
largest leave-one-component-out effects in
Table~\ref{tab:ablation_main}. Removing state refresh affects Memory Awareness
most strongly ($-7.5$\,pp), consistent with the role of the
decay--novelty--enrichment pipeline in distinguishing current from superseded
state. Removing role stabilization affects Profile Understanding most strongly
($-5.5$\,pp), indicating that maintaining profiles depends not only on
retrieving relevant facts but also on preserving responsibility for their
updates.

Removing the Monitor lowers the macro-average by 4.5 points. Under the
implemented topology, Monitor alerts enter the Coordinator's verification and
re-planning path; without those alerts, small role or state inconsistencies can
remain undetected and propagate across later turns. This effect is particularly
relevant in AMemGym's on-policy setting, where one missed intervention can
affect subsequent states.

Removing steering lowers the macro-average by 2.0 points while retaining role
stabilization, state refresh, task-switch correction, memory operations, and
monitoring. This row therefore measures the marginal contribution of spectral
steering within the complete ATLAS pipeline. It should not be interpreted as a
standalone comparison between steering and role stabilization.

\section{Turn-Wise Drift Closure over Long-Horizon Conversations}
\label{app:multidrift_closure}

\begin{figure*}[t]
    \centering
    \includegraphics[width=0.94\textwidth]{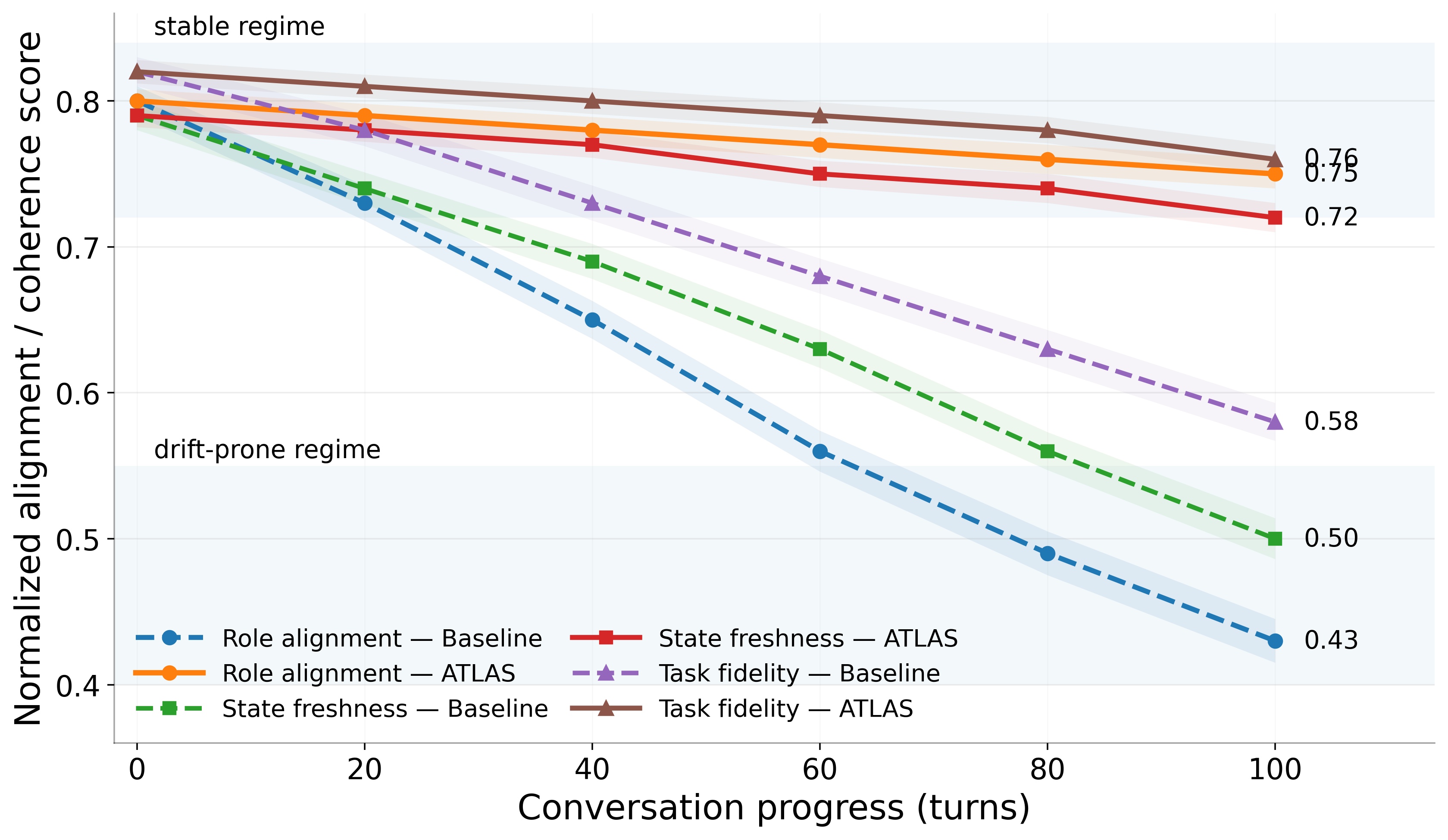}
    \caption{Turn-wise drift closure over a long-horizon multi-agent conversation. We track three coupled coordination dimensions: task fidelity, role alignment, and state freshness, using normalized alignment/coherence scores over the course of the interaction. The baseline progressively drifts across all three dimensions as the horizon increases, with the strongest degradation in role alignment and substantial decay in state freshness and task fidelity. In contrast, ATLAS keeps all three trajectories substantially closer to the stable regime throughout the interaction, indicating that anchor-kernel steering together with adaptive state refresh suppresses the buildup of long-horizon coordination drift. Shaded bands show variability across runs.}
    \label{fig:multidrift_closure_appendix}
\end{figure*}

To directly visualize the phenomenon targeted by ATLAS, we track how coordination quality evolves over the course of a long-horizon interaction. While the main paper reports aggregate improvements in memory performance, role consistency, and robustness, these summary metrics do not explicitly show how drift accumulates turn by turn. Figure~\ref{fig:multidrift_closure_appendix} addresses this by plotting the temporal trajectories of the three coordination dimensions that ATLAS is designed to stabilize: task fidelity, role alignment, and state freshness.

Each curve represents a normalized alignment/coherence score computed at successive points in the interaction. Task fidelity measures how closely the current agent representation remains aligned with the active task specification, role alignment measures consistency with the assigned role-anchor subspace, and state freshness measures how well the agent remains coupled to the latest shared state rather than stale or superseded information. Because these three quantities are modeled jointly by the ATLAS anchor kernel, they provide a compact view of the long-horizon coordination dynamics that the method is intended to control.

The figure shows a clear qualitative difference between the baseline and ATLAS trajectories. In the baseline, all three dimensions degrade as the conversation progresses, showing that long-horizon failures are coupled rather than isolated. Role alignment exhibits the sharpest collapse, which is consistent with the paper's pilot evidence that coherent turns occupy a much higher role-alignment regime than drifted turns. State freshness also declines steadily, reflecting the accumulation of stale or partially superseded memory over long interactions. Task fidelity remains more stable early on, but still decays as the interaction grows longer, indicating that long-range coordination failures eventually pull the system away from the intended task trajectory as well.

In contrast, ATLAS keeps all three curves substantially higher throughout the interaction. This behavior is consistent with the mechanism proposed in the method section. By constructing a joint positive-semidefinite anchor kernel over task, role, and state subspaces, ATLAS repeatedly steers hidden representations toward directions that are simultaneously supported by these sources of coherence. At the same time, the adaptive state-refresh procedure prevents outdated state from dominating later turns. The resulting effect is not merely an improvement in the final response, but an active reduction in the rate at which representational drift accumulates over time.

This diagnostic complements the aggregate experimental results in the main paper. The higher role consistency, stronger long-horizon memory performance, and improved perturbation robustness can be interpreted as downstream consequences of the drift-closure behavior visualized here. In this sense, Figure~\ref{fig:multidrift_closure_appendix} provides a mechanistic bridge between the ATLAS intervention and the task-level gains reported in the main tables.

\section{Secondary Stress Tests}
\label{app:secondary_results}

\subsection{ToolACE}
\label{app:toolace_results}

\begin{table}[t]
\centering
\caption{\textbf{ToolACE results} on 500 multi-step dialogues with Qwen3-8B. AST\,=\,AST Accuracy, AC\,=\,Argument Correctness, MC\,=\,Multi-Step Chain Completion. MemoryArena-Agent and ALARA-Scoped are not applicable (``--'') because ToolACE does not evaluate memory-agent-environment loops or least-privilege scoping. {\scriptsize$\bar\sigma\!=\!1.0$}.}
\label{tab:toolace_app}
\small
\setlength{\tabcolsep}{5pt}
\begin{tabular}{@{}lccc@{}}
\toprule
Method & AST & AC & MC \\
\midrule
Single-Agent       & 74.6 & 61.7 & 42.1 \\
Prompt-Only        & 78.3 & 65.4 & 47.8 \\
Struct-Role        & 80.1 & 67.9 & 50.5 \\
RAG-Mem            & 82.5 & 69.8 & 53.2 \\
A-MEM             & 83.9 & 70.2 & 54.3 \\
MemoryArena-Agent  & --   & --   & --   \\
ActAdd             & 80.8 & 68.3 & 51.0 \\
SEA                & 81.6 & 68.1 & 51.6 \\
ALARA-Scoped       & --   & --   & --   \\
\midrule
\textbf{ATLAS}     & \textbf{88.2} & \textbf{75.8} & \textbf{61.2} \\
\bottomrule
\end{tabular}
\end{table}

Table~\ref{tab:toolace_app} reports the supplementary tool-use stress test. ATLAS improves all three ToolACE metrics over the strongest applicable baseline, A-MEM, with gains of +4.3 on AST, +5.6 on AC, and +6.9 on MC. The disproportionately large gain on MC is significant because this is the strictest metric: every function call in the chain must be correct, and a single incorrect intermediate call invalidates the entire dialogue.

ATLAS's advantage comes from three mechanisms working together. First, the Tool-Executor (in T2/T3) ensures that tool invocation is localized to a dedicated role, reducing the chance that other agents issue unauthorized tool calls. Second, the Memory-Writer reliably persists intermediate tool outputs so that later calls can retrieve them. Third, the Verifier checks whether the proposed next call is consistent with previous tool outputs, catching type mismatches and hallucinated parameter values before they propagate. MemoryArena-Agent is not included because ToolACE does not expose memory-agent-environment metrics, and ALARA-Scoped is not included because ToolACE does not evaluate privilege scoping.

\subsection{ALARA}
\label{app:alara_results}

\begin{table}[t]
\centering
\caption{\textbf{ALARA results} on 115 tasks with Qwen3-8B. TS\,=\,Task Success, PV\,=\,Privilege Violation Rate, OE\,=\,Overexposure Rate. Higher is better for TS; lower is better for PV and OE. MemoryArena-Agent is not applicable (``--''). {\scriptsize$\bar\sigma\!=\!1.2$}.}
\label{tab:alara_app}
\small
\setlength{\tabcolsep}{5pt}
\begin{tabular}{@{}lccc@{}}
\toprule
Method & TS & PV & OE \\
\midrule
Single-Agent       & 54.2 & 18.7 & 23.4 \\
Prompt-Only        & 58.6 & 15.3 & 19.8 \\
Struct-Role        & 63.8 & 12.1 & 16.5 \\
RAG-Mem            & 62.4 & 13.5 & 17.2 \\
A-MEM             & 65.1 & 11.2 & 14.6 \\
MemoryArena-Agent  & --   & --   & --   \\
ActAdd             & 61.7 & 13.8 & 17.9 \\
SEA                & 62.4 & 12.8 & 16.3 \\
ALARA-Scoped       & 67.4 & 9.3  & 12.1 \\
\midrule
\textbf{ATLAS}     & \textbf{72.3} & \textbf{6.8} & \textbf{9.1} \\
\bottomrule
\end{tabular}
\end{table}

Table~\ref{tab:alara_app} reports the least-privilege stress test. ALARA-Scoped improves over generic baselines because it directly enforces declarative context-agent-tool scopes. However, ATLAS further improves task success from 67.4\% to 72.3\% while reducing PV from 9.3\% to 6.8\% and OE from 12.1\% to 9.1\%. This indicates that explicit scoping is helpful but incomplete: scoped access can define what each agent is allowed to see or do, but it does not by itself prevent an agent's latent behavior from drifting toward another role's authority. ATLAS complements scoping by stabilizing the role--task--state representation before output generation, reducing both input-side (PV) and output-side (OE) violations.

\section{Multimodal Stress Test: LIBERO}
\label{app:libero_results}

LIBERO~\cite{liu2023libero} is included as an appendix-only multimodal embodied stress test. It is not a native multi-agent long-conversation benchmark; we use it to test whether the same role--task--state stabilization mechanism transfers to embodied coordination when each manipulation episode is wrapped in a multimodal multi-agent controller.

\paragraph{Protocol.}
LIBERO provides 130 language-conditioned robot manipulation tasks across spatial, object-centric, goal-conditioned, and long-horizon suites. Standard VLA baselines report 90--98\% success on clean evaluation, but recent work shows near-total collapse under perturbations such as displaced objects or rephrased instructions. At each timestep, the environment renders $128\times128$ RGB and depth observations at 10\,Hz. Perception and Depth-Estimator emit structured summaries that State-Tracker fuses into the world model. Planner produces subgoals, Action-Planner maps them to motor templates, and Actuator (OpenVLA-7B) emits 7-DoF continuous actions. We evaluate 3 seeds $\times$ 500 rollouts (10 tasks $\times$ 50 rollouts) per method-stack pair.

Each perturbed episode receives exactly one deterministic perturbation injected between 30\% and 70\% of the expected episode length:

\begin{enumerate}[leftmargin=1.5em,itemsep=2pt]
\item \textbf{Role-confuse}: a distractor message instructs the Coordinator to directly execute an action or ignore the Planner.
\item \textbf{State-delay}: an object position is revised (e.g., 0.15\,m displacement) after Planner has committed to a plan.
\item \textbf{Task-switch}: the language instruction is replaced mid-episode.
\end{enumerate}

\begin{table*}[t]
\centering
\caption{\textbf{LIBERO results} (\%). SR\,=\,Success Rate, PE\,=\,Path Efficiency, SP\,=\,Success under Perturbation. LIBERO is appendix-only. MemoryArena-Agent and ALARA-Scoped are not applicable and marked ``--''. {\scriptsize$\bar\sigma$: M1$\approx$1.8, M2$\approx$1.5, M3$\approx$2.1.}}
\label{tab:libero_app}
\small
\setlength{\tabcolsep}{3.5pt}
\begin{tabular}{@{}l ccc @{\hspace{7pt}} ccc @{\hspace{7pt}} ccc@{}}
\toprule
& \multicolumn{3}{c}{\textbf{M1: Llama+Qwen3VL}}
& \multicolumn{3}{c}{\textbf{M2: Qwen3+Qwen2.5VL}}
& \multicolumn{3}{c}{\textbf{M3: Phi4+Phi4V}} \\
\cmidrule(lr){2-4}\cmidrule(lr){5-7}\cmidrule(lr){8-10}
Method & SR & PE & SP & SR & PE & SP & SR & PE & SP \\
\midrule
Single-Agent       & 72.3 & 64.8 & 18.5 & 75.1 & 67.3 & 20.2 & 68.9 & 61.4 & 15.7 \\
Prompt-Only        & 78.6 & 70.2 & 24.3 & 81.2 & 73.1 & 27.1 & 75.1 & 66.9 & 21.4 \\
Struct-Role        & 81.4 & 73.6 & 29.7 & 84.0 & 76.2 & 32.5 & 78.3 & 70.1 & 26.2 \\
RAG-Mem            & 83.7 & 75.9 & 33.1 & 86.4 & 78.5 & 36.4 & 80.8 & 72.4 & 29.8 \\
A-MEM             & 85.2 & 77.4 & 36.8 & 87.9 & 80.1 & 39.6 & 82.1 & 74.3 & 33.4 \\
MemoryArena-Agent  & --   & --   & --   & --   & --   & --   & --   & --   & -- \\
ActAdd             & 82.1 & 74.3 & 31.5 & 85.0 & 77.1 & 34.8 & 79.3 & 71.0 & 28.2 \\
SEA                & 83.0 & 75.1 & 32.7 & 85.8 & 77.9 & 35.9 & 80.2 & 71.8 & 29.3 \\
ALARA-Scoped       & --   & --   & --   & --   & --   & --   & --   & --   & -- \\
\midrule
\textbf{ATLAS}
& \textbf{89.8} & \textbf{82.6} & \textbf{48.3}
& \textbf{92.1} & \textbf{85.4} & \textbf{51.7}
& \textbf{86.7} & \textbf{79.2} & \textbf{44.1} \\
\bottomrule
\end{tabular}
\end{table*}

ATLAS provides moderate clean-SR gains (+4.2--4.6\,pp) but much larger SP gains (+10.7--12.1\,pp). The main benefit is robustness under coordination stress rather than raw manipulation skill, which is largely determined by the VLA backbone.

\begin{figure*}[t]
    \centering
    \raisebox{-0.0pt}{\includegraphics[width=454.47273pt]{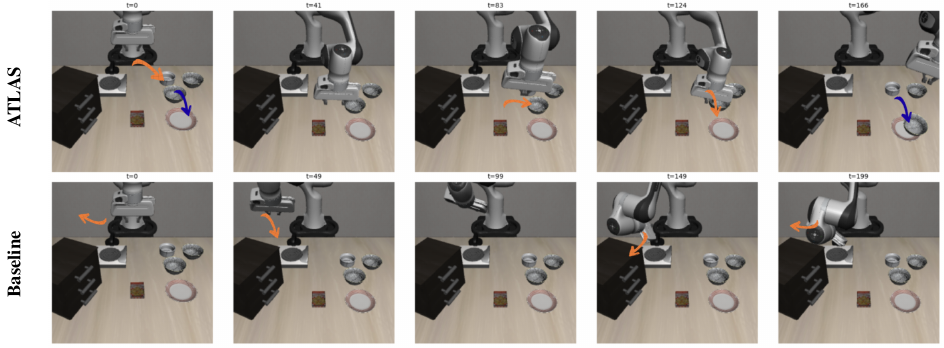}}
    \caption{\textbf{Qualitative LIBERO trajectory under coordination perturbation.}
    We show representative frames from a long-horizon manipulation episode; overlaid arrows mark the intended gripper motion at each frame. The top row shows ATLAS, which maintains the task objective and updated object state across the episode, allowing the controller to recover from intermediate state changes and complete the placement sequence. The bottom row shows the corresponding non-ATLAS trajectory, where coordination drift accumulates over time: the planner and actuator gradually move away from the intended object--target relation, leading to unstable or incomplete execution. The example illustrates the same pattern reflected in Table~\ref{tab:libero_app} and Table~\ref{tab:perturb_app}: ATLAS provides moderate gains on clean success, but substantially larger gains under perturbations that require role-stable planning and fresh state tracking.}
    \label{fig:libero_task0_atlas}
\end{figure*}

Figure~\ref{fig:libero_task0_atlas} provides a qualitative view of the robustness measured quantitatively in Table~\ref{tab:libero_app}. In the ATLAS trajectory, the multimodal controller preserves the intended manipulation objective while updating the world state as the scene evolves. This enables the Planner, State-Tracker, and Actuator to remain coordinated over the full episode. In contrast, the non-ATLAS trajectory shows the characteristic failure mode targeted by our method: small state or role inconsistencies accumulate over time, causing the action sequence to diverge from the intended placement behavior. This qualitative example is consistent with the perturbation breakdown in Table~\ref{tab:perturb_app}, where ATLAS improves most strongly under state-delay perturbations.

\begin{table}[t]
\centering
\caption{\textbf{Perturbation-type breakdown}, LIBERO M2 (\%). SP per perturbation category. {\scriptsize$\bar\sigma\!=\!2.3$}.}
\label{tab:perturb_app}
\small
\setlength{\tabcolsep}{3.5pt}
\begin{tabular}{@{}lccc@{}}
\toprule
Method & \textbf{Role-Conf.} & \textbf{State-Del.} & \textbf{Task-Sw.} \\
\midrule
Prompt-Only       & 34.1 & 12.8 & 34.4 \\
Struct-Role       & 41.3 & 16.5 & 39.7 \\
RAG-Mem           & 44.8 & 20.6 & 43.8 \\
A-MEM            & 48.2 & 24.1 & 46.5 \\
ActAdd            & 42.5 & 19.3 & 42.6 \\
SEA               & 44.6 & 22.3 & 40.8 \\
\midrule
\textbf{ATLAS}    & \textbf{58.9} & \textbf{39.7} & \textbf{56.5} \\
\bottomrule
\end{tabular}
\end{table}

State-delay is the hardest perturbation for all methods (Table~\ref{tab:perturb_app}) because it requires State-Tracker to detect an environmental change after Planner has committed, propagate corrected world state, and trigger re-planning before Actuator executes stale commands. ATLAS improves most strongly on this category (+15.6\,pp over A-MEM), confirming that the decay--novelty--enrichment pipeline effectively incorporates mid-episode state changes. Role-confuse is comparatively easier because Verifier and Monitor catch most routing violations. Task-switch lies between the two because the system must discard stale subgoals and re-orient toward a new objective.

\section{Does ATLAS Require Hidden-State Access?}
\label{sec:hidden_access_ablation}

A central question for ATLAS is whether its gains require direct access to hidden representations, or whether the same task--role--state information could be supplied only through prompts or retrieval. This distinction is important because the main claim of ATLAS is not simply that agents benefit from being reminded of their task, role, and current state. Rather, ATLAS claims that long-horizon coordination failures arise when the internal representation of an agent drifts away from the coupled geometry of these three constraints. If this claim is correct, then exposing the same anchors at the surface-text level should help only partially, while an intervention that corrects the hidden state itself should recover a larger gain.

To test this, we compare five variants that all use the same task, role, and state anchor texts but apply them at different levels of the system. This ablation is run on EverMemBench under the T2 homogeneous twelve-agent setting, which is the main ablation setting in the paper because it contains the full role set, including Tool-Executor and Safety-Checker, while keeping the backbone fixed. The setting therefore isolates whether the benefit comes from the anchor-kernel intervention rather than from heterogeneous model capacity.

\paragraph{Variants.}
\textbf{Prompt ATLAS} inserts the task, role, and state anchors into each agent prompt as explicit natural-language reminders. It receives the same anchor content as ATLAS, but no hidden-state correction is applied. This variant tests whether ATLAS's improvement can be explained by stronger prompt conditioning.

\textbf{Retrieval ATLAS} uses the same anchors only to re-rank retrieved memories, summaries, and state entries before generation. This variant tests whether the benefit comes from better memory selection rather than from representational correction.

\textbf{Random-subspace ATLAS} preserves the same intervention rank, layer, and steering strength as the full method, but replaces the learned task--role--state anchor subspaces with random orthonormal directions. This serves as a control for the possibility that any low-rank perturbation at the intervention layer improves performance.

\textbf{Orthogonalized ATLAS} constructs task, role, and state subspaces from the correct anchors, but orthogonalizes the three bases before steering. This variant keeps hidden-state access and meaningful anchors, but removes the cross-anchor overlap that the full positive-semidefinite kernel is designed to preserve.

\textbf{Hidden ATLAS} is the full method. It combines task, role, and state anchors into a single positive-semidefinite anchor kernel and steers token hidden states toward the kernel's principal eigendirections. Unlike Orthogonalized ATLAS, it preserves directions jointly supported by multiple anchors.

\begin{table}[t]
\centering
\scriptsize
\setlength{\tabcolsep}{4pt}
\renewcommand{\arraystretch}{0.95}
\caption{Hidden-access ablation on EverMemBench under the T2 homogeneous
twelve-agent setting. All variants use the same task--role--state anchors but
apply them through prompting, retrieval ranking, random control subspaces,
orthogonalized hidden steering, or the full ATLAS anchor kernel. Retrieval
ATLAS is the best non-hidden variant. FGR = Fine-Grained Recall, MA = Memory
Awareness, and PU = Profile Understanding.}
\label{tab:hidden_access_ablation}
\begin{tabular}{@{}lcccc@{}}
\toprule
Variant & FGR & MA & PU & Avg \\
\midrule
Prompt ATLAS              & 56.4          & 47.5          & 42.8          & 48.9 \\
Retrieval ATLAS           & 57.2          & 48.6          & 43.6          & 49.8 \\
Random-subspace ATLAS     & 55.0          & 46.8          & 41.7          & 47.8 \\
Orthogonalized ATLAS      & 58.4          & 48.9          & 43.6          & 50.3 \\
Hidden ATLAS              & \textbf{61.7} & \textbf{53.1} & \textbf{47.6} & \textbf{54.1} \\
\midrule
$\Delta$ vs.\ best non-hidden control
                          & $+4.5$        & $+4.5$        & $+4.0$        & $+4.3$ \\
$\Delta$ vs.\ Orthogonalized ATLAS
                          & $+3.3$        & $+4.2$        & $+4.0$        & $+3.8$ \\
\bottomrule
\end{tabular}
\end{table}
\begin{figure*}[t]
    \centering
    \raisebox{-0.0pt}{\includegraphics[width=446.70798pt]{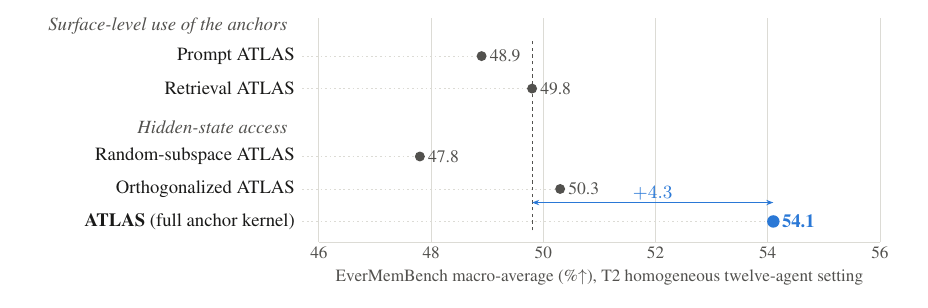}}
    \caption{\textbf{Hidden-access ablation.} Average EverMemBench performance under different ways of using the same task--role--state anchors. Surface-level use through prompting or retrieval improves coordination only partially. Random control subspaces do not recover the gain, and orthogonalizing anchors weakens performance relative to the full positive-semidefinite anchor kernel.}
    \label{fig:hidden_access_ablation}
\end{figure*}

Table~\ref{tab:hidden_access_ablation} and Figure~\ref{fig:hidden_access_ablation} show three consistent trends. First, surface-level access to the anchors is not sufficient. Prompt ATLAS and Retrieval ATLAS improve over ordinary prompt-only or retrieval-only coordination, but they remain below the full hidden-state intervention. This indicates that repeatedly exposing agents to the active task, role boundary, and current state helps, but does not fully prevent internal drift once generation begins.

Second, the random-subspace control performs worse than the structured anchor variants. This rules out a trivial explanation in which ATLAS improves only because it injects a low-rank perturbation at the intervention layer. The direction of the intervention matters: steering must be aligned with task, role, and state geometry rather than with arbitrary directions of the same rank.

Third, Orthogonalized ATLAS performs better than the surface-level controls but worse than Hidden ATLAS. This is the most direct evidence for the anchor-kernel design. Orthogonalized ATLAS has hidden-state access and uses the correct anchor texts, but it removes overlap among task, role, and state directions. The full method preserves this overlap, allowing directions supported by multiple anchors to receive larger effective eigenvalues. The resulting gap shows that the coupling among task intent, role authority, and shared state is not noise to be removed; it is useful coordination structure.

The gains are largest on Memory Awareness and Profile Understanding. This pattern is consistent with the intended mechanism. Memory Awareness requires distinguishing current state from superseded state, while Profile Understanding requires the correct agent role to maintain and update persistent attributes over long interaction horizons. Both settings are sensitive to coupled role--state failures: a memory entry may be available in the context, but the system can still fail if the wrong agent updates it, if the Executor uses an outdated version, or if the Coordinator routes the query according to a stale objective. Hidden ATLAS directly corrects the representation before downstream generation, which helps prevent these failures from propagating across turns.

Overall, this ablation supports the central claim that ATLAS requires more than better prompting or retrieval. The same anchors are useful at the surface level, but the largest improvement comes from applying them inside the model as a coupled spectral object. The result also justifies preserving anchor overlap rather than orthogonalizing task, role, and state into independent directions.

\section{Coordination Diagnostics}
\label{app:coord_metrics}

\begin{table*}[t]
\centering
\caption{\textbf{Coordination metrics} on EverMemBench. RC\,=\,Role Consistency (\%$\uparrow$), SF\,=\,State Freshness (\%$\uparrow$), TSR\,=\,Task-Switch Robustness (ratio$\uparrow$), TDA\,=\,Turn-wise Drift Area ($\downarrow$). {\scriptsize$\bar\sigma$: RC$\approx$1.1, SF$\approx$1.3, TSR$\approx$.03, TDA$\approx$.02.}}
\label{tab:coord_app}
\scriptsize
\setlength{\tabcolsep}{2.5pt}
\begin{tabular}{@{}l cccc @{\hspace{7pt}} cccc @{\hspace{7pt}} cccc@{}}
\toprule
& \multicolumn{4}{c}{\textbf{T1: Homogeneous-10}}
& \multicolumn{4}{c}{\textbf{T2: Homogeneous-12}}
& \multicolumn{4}{c}{\textbf{T3: Heterogeneous-12/6M}} \\
\cmidrule(lr){2-5}\cmidrule(lr){6-9}\cmidrule(lr){10-13}
Method & RC & SF & TSR & TDA & RC & SF & TSR & TDA & RC & SF & TSR & TDA \\
\midrule
Prompt-Only       & 71.3 & 52.4 & 0.61 & 0.38 & 73.8 & 54.9 & 0.64 & 0.35 & 68.5 & 49.7 & 0.57 & 0.41 \\
Struct-Role       & 79.6 & 56.1 & 0.67 & 0.31 & 82.1 & 58.7 & 0.70 & 0.28 & 76.9 & 53.4 & 0.63 & 0.34 \\
RAG-Mem           & 76.4 & 62.8 & 0.69 & 0.29 & 78.9 & 65.3 & 0.72 & 0.26 & 73.8 & 59.7 & 0.65 & 0.32 \\
A-MEM            & 78.2 & 67.5 & 0.72 & 0.26 & 80.7 & 70.1 & 0.75 & 0.23 & 75.6 & 64.8 & 0.68 & 0.29 \\
MemArena-Ag.      & 79.1 & 69.0 & 0.73 & 0.24 & 82.0 & 72.4 & 0.77 & 0.21 & 77.2 & 66.1 & 0.70 & 0.27 \\
ActAdd            & 80.5 & 60.1 & 0.66 & 0.30 & 83.0 & 62.7 & 0.69 & 0.27 & 77.8 & 57.3 & 0.62 & 0.33 \\
SEA               & 81.3 & 61.4 & 0.67 & 0.29 & 83.8 & 63.9 & 0.70 & 0.26 & 78.6 & 58.6 & 0.63 & 0.32 \\
\midrule
\textbf{ATLAS}
& \textbf{91.7} & \textbf{78.4} & \textbf{0.84} & \textbf{0.14}
& \textbf{93.5} & \textbf{81.2} & \textbf{0.87} & \textbf{0.11}
& \textbf{89.1} & \textbf{75.6} & \textbf{0.81} & \textbf{0.17} \\
\bottomrule
\end{tabular}
\end{table*}

Table~\ref{tab:coord_app} reports coordination-specific diagnostics that isolate the three failure modes independently of final-answer accuracy. \textbf{Role Consistency (RC)} measures whether each agent stays within its assigned authority, scored by an independent GPT-4o judge. \textbf{State Freshness (SF)} measures whether state-dependent turns use the latest valid entry. \textbf{Task-Switch Robustness (TSR)} measures accuracy recovery in a 10-turn window after an objective change. \textbf{Turn-wise Drift Area (TDA)} measures cumulative time below a coherence threshold of 0.55.

MemoryArena-Agent improves SF relative to A-MEM because its memory-agent-environment loop strengthens persistent maintenance, but it does not close the gap on RC or TDA because it does not directly correct role/task/state representations. Activation-steering baselines improve RC more than retrieval-based methods but remain weaker on SF because they do not maintain evolving state. ATLAS achieves the best score on every metric in all three settings by addressing both failure modes through a single anchor kernel.

\section{Cost and Topology Scaling}
\label{app:cost_topology}

\begin{table}[t]
\centering
\caption{\textbf{Cost per episode}, T2 EverMemBench. Tok.\,=\,total tokens (K), Lat.\,=\,wall-clock latency (s), C-N\,=\,cost-normalized performance. {\scriptsize$\bar\sigma$: Tok.$\approx$6.1, Lat.$\approx$1.3.}}
\label{tab:cost_app}
\small
\setlength{\tabcolsep}{4pt}
\begin{tabular}{@{}lcccc@{}}
\toprule
Method & Tok. & Calls & Lat. & C-N$\uparrow$ \\
\midrule
Single-Agent      & 28.4  & 1.0  & 4.2  & 1.29 \\
Prompt-Only       & 84.7  & 10.3 & 14.6 & 0.49 \\
RAG-Mem           & 107.3 & 14.1 & 19.3 & 0.43 \\
A-MEM            & 118.5 & 16.7 & 22.1 & 0.41 \\
MemArena-Ag.      & 112.6 & 15.4 & 20.8 & 0.44 \\
SEA               & 91.2  & 10.8 & 16.8 & 0.51 \\
\textbf{ATLAS}    & 96.8  & 12.4 & 17.5 & \textbf{0.56} \\
\bottomrule
\end{tabular}
\end{table}

ATLAS is more expensive than a single-agent system but more efficient than the strongest memory baselines (Table~\ref{tab:cost_app}). Compared with A-MEM, ATLAS uses 18\% fewer tokens, 26\% fewer calls, and 21\% lower latency while achieving a higher macro-average. The savings come from reduced re-planning (representational correction prevents cascading failures) and more effective summarization (well-organized state requires fewer correction turns). ATLAS's spectral operations contribute $<$2\% of wall-clock time because they operate on a small $p \times p$ Gram matrix with $p \ll d$.

\begin{table}[t]
\centering
\caption{\textbf{Topology ablation}, EverMemBench with ATLAS. The 12-agent setting corresponds to T2. {\scriptsize$\bar\sigma\!=\!0.8$}.}
\label{tab:topology_app}
\small
\setlength{\tabcolsep}{4pt}
\begin{tabular}{@{}lcccccc@{}}
\toprule
$N$ & FGR & MA & PU & Avg & Tok. & C-N \\
\midrule
1  & 43.5 & 35.1 & 31.2 & 36.6 & 28.4 & 1.29 \\
3  & 54.3 & 45.6 & 40.1 & 46.7 & 58.2 & 0.80 \\
5  & 58.1 & 49.8 & 44.3 & 50.7 & 74.5 & 0.68 \\
10 & 60.8 & 52.4 & 46.5 & 53.2 & 90.1 & 0.59 \\
12 & 61.7 & 53.1 & 47.6 & 54.1 & 96.8 & 0.56 \\
\bottomrule
\end{tabular}
\end{table}

Gains are sublinear: the largest jump is 1$\to$3 agents (+10.1\,pp), which introduces coordination, execution, and memory separation. Moving 10$\to$12 adds Tool-Executor and Safety-Checker for a smaller +0.9\,pp gain. Practical deployments should match topology to latency budget: the 3-agent ATLAS achieves 86\% of 12-agent accuracy at 60\% of token cost.

\section{Pilot Study Details}
\label{app:pilot_study}

We conduct a pilot study to test whether task drift, role drift, and state
staleness arise from a common representational pattern. The study uses 200
ten-agent conversations generated with Qwen3-8B, each exceeding 100 turns and
spanning collaborative planning, debate, information synthesis, and sequential
decision-making, for a total of 14{,}832 annotated turns.

\paragraph{Annotation protocol.}
For each turn, we extract the hidden state at layer $\ell^\star$ and compute
cosine similarity to the task, role, and state anchor subspaces. Failure labels
are assigned using a frontier LLM-as-judge protocol with GPT-5.5 as the primary
evaluator and Claude Opus 4.7 and Gemini 3.1 Pro as independent secondary
evaluators. Each evaluator receives the dialogue trace, active task, assigned
role, current shared state, and role-specific permission boundary, and
classifies the turn along three binary dimensions: task drift (departure from
the active objective), role drift, and state staleness. A turn is labeled
``drifted'' if it exhibits any failure type. The final label is obtained by
majority vote across the three evaluators. On a 500-turn audit subsample, the
three evaluators reach substantial agreement, with Fleiss' $\kappa=0.78$.
Pairwise agreement is $0.86$ for GPT-5.5--Claude Opus 4.7, $0.84$ for
GPT-5.5--Gemini 3.1 Pro, and $0.82$ for Claude Opus 4.7--Gemini 3.1 Pro.
Disagreements concentrate on borderline role-drift cases where an agent
performs a locally helpful action outside its assigned authority.

The anchor subspaces used for this measurement are the same construction
ATLAS steers toward, so alignment and intervention are not independent by
design. Two controls separate them. A matched-rank random subspace at the
same intervention layer recovers only $47.8$ when steered, against $54.1$
for the learned anchors (Appendix~\ref{sec:hidden_access_ablation}), and
removing the learned span from held-out representations costs $7.2$ points
against $0.8$ for a matched-rank random span
(Appendix~\ref{app:additional_validation}). The alignment signal is
therefore specific to the identified geometry rather than an artifact of
measuring distance to the method's own target.

\begin{table}[t]
\centering
\caption{\textbf{Pilot study: alignment and drift correlation.}
200 conversations, 14{,}832 turns.
$\bar{\cos}$\,=\,cosine similarity to the anchor subspace at $\ell^\star$.
$r$\,=\,Pearson correlation between failure indicators.
$\Delta\bar{\cos}$\,=\,recovery vs.\ Prompt-Only.
{\scriptsize$\bar{\sigma}\approx0.02$}.}
\label{tab:pilot_full}
\small
\setlength{\tabcolsep}{4pt}
\begin{tabular}{@{}lcc@{}}
\toprule
\textbf{Measure} & \textbf{Coherent} & \textbf{Drifted} \\
\midrule
$\bar{\cos}$(task)  & 0.82 & 0.45 \\
$\bar{\cos}$(role)  & 0.80 & 0.41 \\
$\bar{\cos}$(state) & 0.79 & 0.44 \\
$\bar{\cos}$(all)   & 0.81 & 0.43 \\
\midrule
\multicolumn{3}{@{}l}{\textbf{Pairwise drift correlation (Pearson $r$)}} \\
\midrule
task $\leftrightarrow$ role  & \multicolumn{2}{c}{0.78} \\
task $\leftrightarrow$ state & \multicolumn{2}{c}{0.74} \\
role $\leftrightarrow$ state & \multicolumn{2}{c}{0.72} \\
\midrule
\multicolumn{3}{@{}l}{\textbf{Alignment recovery
($\Delta\bar{\cos}$)}} \\
\midrule
Struct-Role & \multicolumn{2}{c}{$+0.02$} \\
RAG-Mem     & \multicolumn{2}{c}{$+0.03$} \\
ActAdd      & \multicolumn{2}{c}{$+0.06$} \\
SEA         & \multicolumn{2}{c}{$+0.06$} \\
ATLAS       & \multicolumn{2}{c}{\textbf{$+0.31$}} \\
\bottomrule
\end{tabular}
\end{table}

Mean alignment drops from 0.81 to 0.43 on drifted turns
($p<10^{-6}$, paired bootstrap). The drop is consistent across all three
anchor types (task: $-0.37$, role: $-0.39$, state: $-0.35$), supporting the
hypothesis that the failures reflect a shared representational phenomenon.
Pairwise correlations ($r\geq0.72$) confirm co-occurrence. Overall, 87\% of
failures occur within two turns of the aggregate alignment score crossing below
0.55, suggesting that representational misalignment precedes observable drift.

ATLAS recovers $+0.31$ in alignment---more than $5\times$ the best non-ATLAS
baseline (ActAdd/SEA at $+0.06$). This gap motivates the joint PSD
anchor-kernel design over independent or orthogonalized steering directions.

\paragraph{Illustrative failure.}
At turn 47, the Memory-Writer begins answering user queries directly (role
violation), the Planner references a budget value overridden 12 turns earlier
(state staleness), and the Coordinator routes toward the pre-switch objective
(task drift). Their hidden-state trajectories align: role-subspace cosine
similarity falls from 0.83 to 0.39 between turns 42--47, with task and state
alignment declining in parallel. This illustrates the coupled nature of the
three failure modes.
\section{Additional Method Details}
\label{app:method_details}

\subsection{Intervention Layer Selection}
\label{app:probe}

ATLAS intervenes at a single layer $\ell^\star$ in the middle third of the
model, selected using a linear probe that separates task-consistent from
task-inconsistent continuations~\citep{zou2023representation,li2023iti}. For
each candidate layer $\ell$ in the range $[L/3,2L/3]$, we collect hidden states
from 500 validation conversations and train an $L_2$-regularized logistic
regression ($C=1.0$). The layer with the highest validation AUROC is selected:
$\ell^\star=18$ for Qwen3-8B (AUROC $=0.91$), $\ell^\star=28$ for
Llama-4-Scout ($0.89$), and $\ell^\star=24$ for Phi-4 ($0.90$). All selected
layers lie near the midpoint of the eligible range, consistent with mid-layer
representations capturing abstract semantic properties. On Qwen3-8B, the
macro-average varies by less than $0.8$\,pp across $\ell^\star\pm2$.

\subsection{Low-Rank Kernel Computation}
\label{app:kernel_computation}

The anchor kernel is never materialized as a dense $d\times d$ matrix. We use the factorization $\mat{K}^{(t)}=\mat{F}^{(t)}(\mat{F}^{(t)})^\top$ where $\mat{F}^{(t)}\in\mathbb{R}^{d\times p}$ with $p\leq60$ for $d=4096$. The Gram matrix $\mat{G}^{(t)}=(\mat{F}^{(t)})^\top\mat{F}^{(t)}\in\mathbb{R}^{p\times p}$ has cost $O(dp^2)$ to form and $O(p^3)$ to eigendecompose. Eigenvectors of $\mat{K}^{(t)}$ are recovered by lifting: $\vect{v}_j=\mat{F}^{(t)}\vect{z}_j/\|\mat{F}^{(t)}\vect{z}_j\|$. Total per-turn cost is $O(dp^2+p^3)$, giving a $\sim$$4{,}500\times$ speedup over dense $O(d^3)$. The fixed task/role blocks are cached across turns; only state columns are updated.

\subsection{Why Anchor Overlap Matters}
\label{app:overlap}

If a direction $\vect{u}$ lies in both the task and state subspaces, its effective eigenvalue receives additive contributions: $\lambda\approx w_T\alpha_T^2 + w_S\sigma_j\alpha_S^2$. Overlap makes jointly relevant directions more prominent during steering. Orthogonalizing anchors removes this coupling, treating task, role, and state as independent constraints. On Qwen3-8B EverMemBench, orthogonalized ATLAS achieves 50.3\% vs.\ 54.1\% for the full method ($-$3.8\,pp), concentrated in MA ($-$4.2) and PU ($-$3.9), confirming that anchor overlap is empirically important.

\subsection{Steering Analysis}
\label{app:steering_analysis}

The coherence score
\[
\mathcal{C}(\vect{h})
=
\frac{\vect{h}^{\top}\mat{K}\vect{h}}{\|\vect{h}\|^2}
\]
and spectral variance
\[
\mathcal{V}(\vect{h})
=
\frac{\vect{h}^{\top}\mat{K}^2\vect{h}}{\|\vect{h}\|^2}
-
\mathcal{C}(\vect{h})^2
\]
govern ATLAS steering.

\begin{theorem}[Continuous coherence monotonicity]
\label{thm:monotone}
For $\vect{h}\neq\vect{0}$ evolving according to
$d\vect{h}/d\alpha=\mat{K}\vect{h}$ with
$\mat{K}\succeq0$,
\[
\frac{d\mathcal{C}}{d\alpha}
=
2\mathcal{V}(\vect{h})
\geq0.
\]
\end{theorem}

\begin{proof}
Differentiating
$\mathcal{C}(\vect{h})
=\vect{h}^{\top}\mat{K}\vect{h}/(\vect{h}^{\top}\vect{h})$
along $d\vect{h}/d\alpha=\mat{K}\vect{h}$ gives
\[
\frac{d\mathcal{C}}{d\alpha}
=
2\left(
\frac{\vect{h}^{\top}\mat{K}^2\vect{h}}{\|\vect{h}\|^2}
-
\left(
\frac{\vect{h}^{\top}\mat{K}\vect{h}}{\|\vect{h}\|^2}
\right)^2
\right)
=
2\mathcal{V}(\vect{h}).
\]
Extending the eigendecomposition of $\mat{K}$ to include its null
space, let
$p_j=(\vect{v}_j^\top\vect{h})^2/\|\vect{h}\|^2$.
Then $\sum_jp_j=1$, and $\mathcal{V}(\vect{h})$ is the variance of
the eigenvalues $\{\lambda_j\}$ under $\{p_j\}$; hence it is
nonnegative.
\end{proof}

\begin{theorem}[Discrete coherence monotonicity]
\label{thm:discrete_monotone}
For the steering update in Eq.~\eqref{eq:steer_main}, with
$a=\alpha_{\mathrm{eff}}\geq0$,
\[
\mathcal{C}\!\left((\mat{I}+a\mat{K})\vect{h}\right)
\geq
\mathcal{C}(\vect{h}).
\]
\end{theorem}

\begin{proof}
Using the full eigenbasis of $\mat{K}$ and the probabilities $p_j$
defined above,
\[
\mathcal{C}\!\left((\mat{I}+a\mat{K})\vect{h}\right)
=
\frac{
\sum_j\lambda_j(1+a\lambda_j)^2p_j
}{
\sum_j(1+a\lambda_j)^2p_j
}.
\]
Because $g(\lambda)=(1+a\lambda)^2$ is non-decreasing for
$a\geq0$ and $\lambda\geq0$, Chebyshev's association inequality
gives
\[
\operatorname{Cov}_{p}\!\left(\lambda,g(\lambda)\right)\geq0.
\]
Consequently,
\[
\mathcal{C}\!\left((\mat{I}+a\mat{K})\vect{h}\right)
-
\mathcal{C}(\vect{h})
=
\frac{
\operatorname{Cov}_{p}\!\left(\lambda,g(\lambda)\right)
}{
\mathbb{E}_{p}[g(\lambda)]
}
\geq0.
\]
\end{proof}

Thus, coherence is non-decreasing under both the continuous dynamics
and the discrete update. Its continuous increase is governed by
spectral variance and can vanish when $\vect{h}$ is concentrated in a
single eigenspace, including the null space. The cap
$\alpha_{\mathrm{eff}}=\min(\alpha,\xi/\lambda_1)$ with $\xi=0.15$
is not required for monotonicity; it limits the largest
per-direction correction.

\subsection{Adaptive State-Memory Update}
\label{app:memory_update}

The state component evolves via three steps. \textbf{Decay}: existing eigenvalues are multiplied by $\gamma=0.95$, so every retained direction loses half its current importance in
$\sim$14 turns. \textbf{Novelty}: the turn-level representation is projected away from the current state subspace; if the residual norm fraction $\nu$ exceeds threshold $\rho=0.10$, the normalized residual is treated as a novel state direction. \textbf{Enrichment}: the novel direction is appended with initial importance $\beta_0=1.0$; if rank exceeds $k_S^{\max}=30$, the weakest direction is pruned. Re-orthonormalization via modified Gram--Schmidt is applied every $\Delta=5$ turns. Multi-agent updates within a turn are appended sequentially; the Gram-matrix eigendecomposition is invariant to column ordering.

\subsection{State-Enrichment Stability}
\label{app:state_stability}

Each enrichment adds a rank-one PSD perturbation $w_S\beta_0\hat{\vect{m}}\hat{\vect{m}}^\top$.

\begin{theorem}[Bounded spectral shift]
\label{thm:interlace}
Each eigenvalue shifts by at most $w_S\beta_0$; the trace increases by exactly $w_S\beta_0$.
\end{theorem}

\begin{proof}
The perturbation has operator norm $w_S\beta_0$. Weyl's inequality bounds individual eigenvalue shifts. Trace increase is $w_S\beta_0\|\hat{\vect{m}}\|^2=w_S\beta_0$.
\end{proof}

With $w_S=1/3$ and $\beta_0=1.0$, the maximum per-step eigenvalue change is $\sim$0.33. Decay offsets most trace growth, and pruning bounds rank, so the kernel remains stable over episodes of several hundred turns.

\section{Additional Validation, Robustness, and Scaling}
\label{app:additional_validation}

This section provides additional controls for the principal design
choices in ATLAS.  We examine component complementarity, scaling to
a substantially larger dense backbone, evaluator reliability and
interventional evidence, robustness to anchor specification and
hyperparameters, and the implementation of ATLAS in heterogeneous
model teams.

\subsection{Controlled Evaluation Setting}
\label{app:additional_controls}

The gains reported below do not arise from an ATLAS-specific agent
topology or evaluation protocol.  Within each experimental setting,
all methods use the same agent roles, episode-level 128K context
budget, top-$k{=}10$ retrieval depth, 20-turn summarization interval,
200-turn episode limit, greedy decoding, random seeds, and evaluation
procedure.  The topology is fixed across methods as described in
Appendix~A1.1, while Appendix~A1.4 specifies the shared experimental
settings and validation procedure.  ATLAS, ActAdd, and SEA receive
the same held-out validation budget for intervention-layer and
steering-strength selection; memory baselines are tuned on the same
validation set.

Benchmark-specific methods are used only where their objectives and
metrics remain compatible.  MemoryArena-Agent is evaluated on
EverMemBench and AMemGym, where its memory-agent-environment loop can
be adapted without changing the evaluation objective.
ALARA-Scoped is used only for the least-privilege ALARA evaluation.
LIBERO remains an appendix-only embodied stress test, with its task
definitions, observations, action space, and success criterion
preserved.  This avoids attributing differences to incompatible
baseline adaptations.

The controlled hidden-access experiment in Appendix~A5 provides an
additional system-level control.  With the surrounding topology and
memory pipeline fixed, Prompt ATLAS, Retrieval ATLAS,
Random-subspace ATLAS, and Orthogonalized ATLAS obtain averages of
48.9, 49.8, 47.8, and 50.3, respectively, compared with 54.1 for
full Hidden ATLAS.  The 3.8-point gap over the strongest controlled
variant isolates the benefit of hidden-state intervention while
preserving task--role--state overlap.

\subsection{Component Complementarity and Kernel Composition}
\label{app:kernel_composition}

Role stabilization and spectral steering operate at different stages
of ATLAS and are therefore not interchangeable.  The agent-specific
role basis $U_R^{(a)}$ is an input to the joint task--role--state
kernel $K^{(t)}$ in Eq.~(2), whereas spectral steering in Eq.~(8)
is the operation that applies this kernel to token hidden states.
The role basis specifies which representational directions encode
the agent's assigned authority; steering performs the corresponding
hidden-state correction.

The leave-one-out ablation in \S\ref{ssec:ablations_cost} already separates these
effects.  Removing steering while retaining role prompting, state
refresh, task-switch correction, and monitoring reduces the
EverMemBench T2 average from 54.1 to 52.1 ($-2.0$).  Conversely,
removing the role anchor while retaining steering reduces performance
to 49.1 ($-5.0$), and removing adaptive state refresh reduces it to
48.9 ($-5.2$).  Thus, the role anchor provides role-specific
geometry, while steering converts that geometry into an internal
representational correction.

We further vary which anchors enter the kernel while leaving all
other components unchanged.  For reduced kernels, the omitted anchor
term is zeroed while the retained terms use the same weights as in
the full configuration; the remaining weights are not renormalized.
Consequently, the $T{+}S$ configuration is identical to the
``w/o role stabilization'' ablation.

\begin{table}[t]
\centering
\small
\setlength{\tabcolsep}{5pt}
\begin{tabular}{@{}lc@{}}
\toprule
Kernel configuration & Avg. \\
\midrule
Task only ($T$)       & 46.8 \\
Role only ($R$)       & 47.6 \\
State only ($S$)      & 48.3 \\
$T+R$                 & 50.4 \\
$T+S$                 & 49.1 \\
$R+S$                 & 51.6 \\
\textbf{Full ($T+R+S$)} & \textbf{54.1} \\
\bottomrule
\end{tabular}
\caption{\textbf{Kernel-composition ablation on T2
EverMemBench.} Only the anchors included in the steering kernel are
varied; the remaining ATLAS pipeline is held fixed.  Every singleton
and pairwise kernel underperforms the full coupled kernel.}
\label{tab:kernel_composition}
\end{table}

Every singleton and pairwise configuration underperforms the full
kernel.  Moreover, the full $T{+}R{+}S$ configuration is the only
kernel composition that exceeds the no-steering variant (52.1).
Together with the 50.3 result for Orthogonalized ATLAS in
Appendix~A5, these results show that the improvement is not produced
by several independent modules being stacked: it depends on applying
steering to a joint PSD kernel that preserves useful overlap among
task, role, and state directions.

\subsection{Scaling to a Larger Dense Backbone}
\label{app:large_model_scaling}

The main evaluation already includes Llama-4-Scout-17B-16E, a
mixture-of-experts model with 17B active parameters, 16 experts, and
109B total parameters, as well as a heterogeneous six-model T3
configuration.  This nevertheless leaves open performance on a
dense model substantially above 20B parameters.

ATLAS's additional spectral cost depends primarily on the hidden
dimension $d$ and combined anchor rank $p$, rather than directly on
the backbone's total parameter count.  As described in
Appendix~A9.2, the dense $d\times d$ kernel is never materialized.
The intervention operates through the $p\times p$ Gram matrix with
cost $O(dp^2+p^3)$.  The underlying forward-pass cost naturally
grows with model size, but the ATLAS intervention requires no
retraining and uses the same layer-selection, anchor-construction,
and steering procedure.

We therefore evaluate Qwen2.5-72B-Instruct (72.7B parameters,
80 layers) under the T2 EverMemBench protocol.  The experiment uses
the same twelve-agent topology, 128K episode-level context budget,
top-$k{=}10$ retrieval, decoding procedure, and evaluation metrics,
with means over three seeds.

\begin{table}[t]
\centering
\small
\setlength{\tabcolsep}{3.5pt}
\resizebox{\columnwidth}{!}{
\begin{tabular}{@{}lcccc@{}}
\toprule
Method & FGR & MA & PU & Avg. \\
\midrule
A-MEM             & 63.4 & 54.1 & 48.5 & 55.3 \\
MemoryArena-Agent  & 64.1 & 55.0 & 49.5 & 56.2 \\
\textbf{ATLAS}     & \textbf{69.0} & \textbf{59.8} &
                     \textbf{54.2} & \textbf{61.0} \\
\bottomrule
\end{tabular}}
\caption{\textbf{Dense-backbone scaling on T2 EverMemBench using
Qwen2.5-72B-Instruct.} All methods use the same topology, context
budget, retrieval settings, decoding procedure, and evaluation.}
\label{tab:qwen72b}
\end{table}

ATLAS obtains a macro-average of 61.0, improving over
MemoryArena-Agent, the strongest memory baseline, by 4.8 points.
The magnitude of the gain is therefore comparable to that observed
with the smaller primary backbones, indicating that the
anchor-kernel intervention remains effective at substantially larger
dense-model scale.

\subsection{Evaluator Reliability and Interventional Evidence}
\label{app:judge_validation}

The headline EverMemBench and AMemGym results do not depend on the
multi-model judges used in the pilot drift analysis.  EverMemBench
Fine-Grained Recall is exact-match against gold answers, while
Memory Awareness and Profile Understanding use the benchmark's
official semantic-matching scorer identically for every method.
AMemGym write, read, and utilization failures are scored against its
oracle traces.  The GPT-5.5/Claude/Gemini judge ensemble is used for
the diagnostic coupling study in Appendix~A8, while GPT-4o is used
separately for the Role-Consistency diagnostic.

Appendix~A8 reports pairwise agreement of 0.82--0.86 among the three
pilot evaluators and Fleiss' $\kappa=0.78$.  Disagreements are
concentrated in borderline cases in which an agent performs a useful
action while exceeding its assigned authority.  We additionally
audit 200 randomly sampled turns.  Two independent annotators with
graduate-level research training label the turns using the same
definitions of task drift, role drift, and state staleness while
remaining blinded to the model-judge decisions.  Human consensus is
obtained after the two independent annotations; for comparison with
the GPT-4o Role-Consistency judge, the corresponding human
role-consistency label is used.

\begin{table}[t]
\centering
\small
\setlength{\tabcolsep}{3.5pt}
\resizebox{\columnwidth}{!}{
\begin{tabular}{@{}lccc@{}}
\toprule
Comparison & Turns & Agreement & Cohen's $\kappa$ \\
\midrule
Human 1 vs.\ Human 2
    & 200 & \textbf{91.0\%} & \textbf{0.82} \\
Human consensus vs.\ model majority
    & 200 & \textbf{89.5\%} & \textbf{0.79} \\
Human consensus vs.\ GPT-4o RC judge
    & 200 & \textbf{88.5\%} & \textbf{0.76} \\
\bottomrule
\end{tabular}}
\caption{\textbf{Human audit of judge-based diagnostics.} Human
annotations are produced independently and without access to the
model-judge decisions.}
\label{tab:human_audit}
\end{table}

The human audit shows close correspondence between independent human
decisions and the judge-based labels, with residual disagreements
again dominated by useful-but-role-exceeding actions.

The evidence for the proposed anchor geometry also extends beyond
correlation.  In the pilot study, aggregate anchor alignment falls
from 0.81 on coherent turns to 0.43 on drifted turns, and 87\% of
observed failures occur within two turns after aggregate alignment
crosses below 0.55.  This establishes temporal precedence in the
observed traces.  Appendix~A5 further provides intervention-based
controls: matched-rank random-subspace steering reaches only 47.8,
while orthogonalizing the correct task--role--state anchors reduces
the average from 54.1 to 50.3.

To probe the functional role of the anchor span more directly, we
remove either the learned joint task--role--state span or a
matched-rank random span at the same intervention layer and rank on
held-out coherent episodes.

\begin{table}[t]
\centering
\small
\setlength{\tabcolsep}{5pt}
\begin{tabular}{@{}lcc@{}}
\toprule
Hidden-state condition & Avg. & Change \\
\midrule
No removal                    & \textbf{54.1} & -- \\
Matched-rank random removal   & 53.3 & $-0.8$ \\
Remove task--role--state span & 46.9 & $-7.2$ \\
\bottomrule
\end{tabular}
\caption{\textbf{Anchor-removal intervention on T2
EverMemBench.} Removing the learned task--role--state span produces
a substantially larger degradation than removing a random span of
the same rank.}
\label{tab:anchor_removal}
\end{table}

Removing the learned anchor span produces a 7.2-point degradation,
compared with 0.8 points under the matched-rank random control.
Together with the temporal and random-steering analyses, this result
supports a functional role for the proposed task--role--state
geometry.  We therefore interpret the shared-cause account as a
hypothesis supported by diagnostic, temporal, and interventional
evidence rather than as a causal claim established from correlation
alone.

\subsection{Anchor Quality and Hyperparameter Robustness}
\label{app:anchor_robustness}

Because anchor subspaces are constructed from natural-language task,
role, and state descriptions, their quality is a potential source of
sensitivity.  We therefore perturb the anchor descriptions while
holding all other T2 EverMemBench settings fixed.

\begin{table}[t]
\centering
\small
\setlength{\tabcolsep}{5pt}
\begin{tabular}{@{}lcc@{}}
\toprule
Anchor condition & Avg. & $\Delta$ \\
\midrule
Original anchors          & \textbf{54.1} & -- \\
Paraphrased, 3 sets       & 53.8 & $-0.3$ \\
75\% length               & 53.6 & $-0.5$ \\
50\% length               & 52.4 & $-1.7$ \\
25\% length               & 49.2 & $-4.9$ \\
30\% lexical noise        & 50.1 & $-4.0$ \\
Contradiction injection   & 46.5 & $-7.6$ \\
Role-swap corruption      & 44.4 & $-9.7$ \\
\bottomrule
\end{tabular}
\caption{\textbf{Robustness to anchor quality on T2
EverMemBench.} Semantics-preserving paraphrases and moderate
shortening have little effect, whereas severe semantic corruption
produces predictable degradation.}
\label{tab:anchor_quality}
\end{table}

ATLAS is stable to paraphrasing and moderate shortening but degrades
as anchor semantics are progressively damaged.  Contradiction
injection reduces the average to 46.5, close to RAG-Mem (46.6 in
T2), while role-swap corruption reaches 44.4, close to Struct-Role
(44.0).  These conditions provide an empirical lower-bound
characterization when the anchor specification itself is unreliable.

We also vary each principal hyperparameter independently while
holding the others at their default values:
$\alpha=0.15$, $\gamma=0.95$, $\rho=0.10$, $\beta_0=1.0$,
$\tau=0.90$, and $\xi=0.15$.  The default configuration obtains
54.1 average.

\begin{table}[t]
\centering
\scriptsize
\setlength{\tabcolsep}{3pt}
\resizebox{\columnwidth}{!}{
\begin{tabular}{@{}lll@{}}
\toprule
Parameter & Alternative settings & EverMemBench Avg. \\
\midrule
$\alpha$  & 0.05 / 0.10 / 0.20 & 51.4 / 53.2 / 53.6 \\
$\gamma$  & 0.90 / 0.98        & 52.7 / 54.3 \\
$\rho$    & 0.05 / 0.15        & 53.0 / 53.8 \\
$\beta_0$ & 0.5 / 1.5          & 52.9 / 54.0 \\
$\tau$    & 0.85 / 0.95        & 53.4 / 54.2 \\
$\xi$     & 0.10 / 0.20        & 53.5 / 53.9 \\
\bottomrule
\end{tabular}}
\caption{\textbf{One-at-a-time hyperparameter sensitivity on T2
EverMemBench.} All unlisted hyperparameters remain at their default
values.}
\label{tab:hyperparameter_sensitivity}
\end{table}

Most neighboring configurations remain close to the default, and
the largest degradation occurs under very weak steering
($\alpha=0.05$).  The intervention therefore does not rely on a
single narrowly tuned configuration.  The search ranges for
$\alpha,\gamma,\rho,$ and $\beta_0$ and their selected values are
reported in Appendix~A1.4; the variance threshold $\tau$ and
steering cap $\xi$ are included here to make the complete default
configuration explicit.

\subsection{Heterogeneous-Model Instantiation and Layer Selection}
\label{app:heterogeneous_details}

ATLAS does not require agents to share a hidden dimension or a common
representation coordinate system.  For each agent $a$, task, role,
and state anchors are constructed using that agent's own frozen
backbone.  Its kernel $K_a^{(t)}$ and spectral intervention therefore
remain entirely within that model's native hidden space.  No hidden
state, anchor basis, or spectral direction is projected between
different model families.  Cross-model coordination occurs through
natural-language messages and the shared state, as in the
corresponding baselines.

The T3 configuration distributes the twelve roles across six model
families: Llama-4-Scout-17B-16E, Qwen3-8B,
Qwen2.5-7B-Instruct, Phi-4-14B, Gemma-2-9B-Instruct, and
Mistral-7B-Instruct.  The role-to-model assignments are given in
Appendix~A1.2.  Intervention layers are selected separately within
each backbone using the same middle-third linear-probe procedure
described in Appendix~A9.1.  For the primary reported backbones,
the selected layers are $\ell^\star=18$ for Qwen3-8B
(AUROC $=0.91$), $\ell^\star=28$ for Llama-4-Scout
($0.89$), and $\ell^\star=24$ for Phi-4 ($0.90$).
On Qwen3-8B, moving the intervention by
$\ell^\star\pm2$ changes the macro-average by less than 0.8 points.
ActAdd and SEA receive the same per-backbone layer-selection and
steering-strength validation budget.

To determine whether ATLAS benefits disproportionately from
per-backbone layer selection, we additionally evaluate all three
activation-steering methods using the same relative-depth rule,
$\ell=\lfloor0.5L\rfloor$, for every backbone.

\begin{table}[t]
\centering
\small
\setlength{\tabcolsep}{4pt}
\begin{tabular}{@{}lccc@{}}
\toprule
Method & Midpoint & Per-backbone $\ell^\star$ & Gain \\
\midrule
ActAdd & 40.4 & 41.0 & $+0.6$ \\
SEA    & 41.3 & 42.0 & $+0.7$ \\
ATLAS  & 47.9 & \textbf{49.2} & $+1.3$ \\
\bottomrule
\end{tabular}
\caption{\textbf{Layer-selection control in the heterogeneous T3
setting.} ``Midpoint'' applies
$\ell=\lfloor0.5L\rfloor$ uniformly to every backbone, whereas the
second column uses the per-backbone probe-selection procedure.}
\label{tab:layer_fairness}
\end{table}

Per-backbone selection improves each activation-steering method
modestly, while ATLAS remains substantially ahead under the common
midpoint rule.  The heterogeneous gain therefore cannot be explained
by preferential layer tuning.

Finally, ATLAS requires access to intermediate hidden states and is
therefore most directly applicable to open-weight or otherwise
controllable inference stacks.  Stronger coordination does not by
itself guarantee factual correctness, safety, or optimal tool use,
and increased agent capability may also facilitate harmful
workflows.  In safety-sensitive deployments, the representational
intervention should therefore be combined with privilege controls,
logging, human oversight, and external safety checks, consistent
with the limitations and ethics discussion in the main paper.

\clearpage

\begin{algorithm}[t]
\small
\caption{ATLAS}
\label{alg:atlas}
\begin{algorithmic}[1]
\Require Frozen LLM, layer $\ell^\star$; anchors $T$, $\{R^{(a)}\}$, $S^{(0)}$; weights $w_T,w_R,w_S$; gain $\alpha$, cap $\xi$; decay $\gamma$, threshold $\rho$, importance $\beta_0$, max rank $k_S^{\max}$, period $\Delta$
\Statex \textit{Offline: anchor subspaces}
\For{$c \in \{T\} \cup \{R^{(a)}\}_a \cup \{S^{(0)}\}$}
  \State Collect layer-$\ell^\star$ states $\mat{X}_c$
  \State $\mat{U}_c \gets$ top-$k_c$ left singular vectors of $\mat{X}_c$ covering variance $\tau$ \Comment{Eq.~\ref{eq:basis_main}}
\EndFor
\Statex \textit{Online inference}
\For{turn $t=1,2,\ldots$, active agent $a$}
  \State $\mat{F}^{(t)} \gets \big[\sqrt{w_T}\mat{U}_T,\, \sqrt{w_R}\mat{U}^{(a)}_R,$
  \Statex \hspace{4.2em} $\sqrt{w_S}\mat{V}^{(t)}_S (\mat{\Sigma}^{(t)}_S)^{1/2}\big]$ \Comment{Eq.~\ref{eq:F_main}}
  \State Eigendecompose $(\mat{F}^{(t)})^\top \mat{F}^{(t)}$; lift to $\{(\lambda_j,\vect{v}_j)\}_{j=1}^{r}$ of $\mat{K}^{(t)}$
  \State $\alpha_{\mathrm{eff}} \gets \min(\alpha, \xi/\lambda_1)$ \Comment{Eq.~\ref{eq:alpha_eff_main}}
  \For{token $i=1,\ldots,L_t$}
    \State $\widetilde{\vect{h}}^{(a,i)}_t \gets \vect{h}^{(a,i)}_t \,+$
    \Statex \hspace{4.2em} $\alpha_{\mathrm{eff}} \sum_{j} \lambda_j (\vect{v}_j^\top \vect{h}^{(a,i)}_t) \vect{v}_j$ \Comment{Eq.~\ref{eq:steer_expanded_main}}
  \EndFor
  \Statex \textit{\quad State memory update}
  \State $\bar{\sigma}^{(t)}_j \gets \gamma \sigma^{(t)}_j$, $j=1,\ldots,k_S$ \Comment{Eq.~\ref{eq:decay_main}}
  \State $\bar{\widetilde{\vect{h}}}^{(a)}_t \gets \frac{1}{L_t}\sum_i \widetilde{\vect{h}}^{(a,i)}_t$ \Comment{Eq.~\ref{eq:turn_mean_main}}
  \State $\vect{m} \gets (\mat{I} - \mat{V}^{(t)}_S (\mat{V}^{(t)}_S)^\top) \bar{\widetilde{\vect{h}}}^{(a)}_t$
  \State $\nu \gets \|\vect{m}\|^2 / \|\bar{\widetilde{\vect{h}}}^{(a)}_t\|^2$ \Comment{Eq.~\ref{eq:novelty_main}}
  \If{$\nu > \rho$} \Comment{enrich}
    \State $\mat{V}^{(t+1)}_S \gets [\mat{V}^{(t)}_S,\, \vect{m}/\|\vect{m}\|]$
    \State $\mat{\Sigma}^{(t+1)}_S \gets \mathrm{diag}(\bar{\sigma}^{(t)}_1,\ldots,\bar{\sigma}^{(t)}_{k_S}, \beta_0)$
    \If{rank $> k_S^{\max}$}
      \State Drop smallest-eigenvalue direction
    \EndIf
  \Else \Comment{carry forward}
    \State $\mat{V}^{(t+1)}_S \gets \mat{V}^{(t)}_S$
    \State $\mat{\Sigma}^{(t+1)}_S \gets \mathrm{diag}(\bar{\sigma}^{(t)}_1,\ldots,\bar{\sigma}^{(t)}_{k_S})$
  \EndIf
  \If{$t \bmod \Delta = 0$}
    \State Re-orthonormalize $\mat{V}^{(t+1)}_S$
  \EndIf
\EndFor
\end{algorithmic}
\end{algorithm}

%% file: custom.bib
@misc{jiayang2026amemgym,
      title={AMemGym: Interactive Memory Benchmarking for Assistants in Long-Horizon Conversations}, 
      author={Cheng Jiayang and Dongyu Ru and Lin Qiu and Yiyang Li and Xuezhi Cao and Yangqiu Song and Xunliang Cai},
      year={2026},
      eprint={2603.01966},
      archivePrefix={arXiv},
      primaryClass={cs.CL},
      url={https://arxiv.org/abs/2603.01966}, 
}

@inproceedings{liu2023libero,
  title     = {{LIBERO}: Benchmarking Knowledge Transfer for Lifelong Robot Learning},
  author    = {Liu, Bo and Zhu, Yifeng and Gao, Chongkai and Feng, Yihao and Liu, Qiang and Zhu, Yuke and Stone, Peter},
  booktitle = {Advances in Neural Information Processing Systems},
  volume    = {36},
  year      = {2023},
  url       = {https://proceedings.neurips.cc/paper_files/paper/2023/hash/8c3c666820ea055a77726d66fc7d447f-Abstract-Datasets_and_Benchmarks.html},
  doi       = {10.52202/075280-1939}
}

@misc{agostino2026alara,
      title={Herding CATs: ALARA for Agent Harness Engineering in Portable Composable Multi-Agent Teams}, 
      author={Christopher J. Agostino and Nayan D'Souza},
      year={2026},
      eprint={2603.20380},
      archivePrefix={arXiv},
      primaryClass={cs.MA},
      url={https://arxiv.org/abs/2603.20380}, 
}

@inproceedings{patil2025bfcl,
  title     = {The {Berkeley} Function Calling Leaderboard ({BFCL}): From Tool Use to Agentic Evaluation of Large Language Models},
  author    = {Patil, Shishir G. and Mao, Huanzhi and Yan, Fanjia and Ji, Charlie Cheng-Jie and Suresh, Vishnu and Stoica, Ion and Gonzalez, Joseph E.},
  booktitle = {Proceedings of the International Conference on Machine Learning},
  year      = {2025},
  url       = {https://proceedings.mlr.press/v267/patil25a.html}
}

@inproceedings{xu2025amem,
  title     = {{A-MEM}: Agentic Memory for {LLM} Agents},
  author    = {Xu, Wujiang and Liang, Zujie and Mei, Kai and Gao, Hang and Tan, Juntao and Zhang, Yongfeng},
  booktitle = {Advances in Neural Information Processing Systems},
  year      = {2025},
  doi       = {10.52202/085713-0593}
}

@misc{turner2024actadd,
  title     = {Steering Language Models With Activation Engineering},
  author    = {Turner, Alexander Matt and Thiergart, Lisa and Leech, Gavin and Udell, David and Vazquez, Juan J. and Mini, Ulisse and MacDiarmid, Monte},
  year      = {2023},
  eprint        = {2308.10248},
  archivePrefix = {arXiv},
  url           = {https://arxiv.org/abs/2308.10248}
}

@inproceedings{qiu2024sea,
  title     = {Spectral Editing of Activations for Large Language Model Alignment},
  author    = {Qiu, Yifu and Zhao, Zheng and Ziser, Yftah and Korhonen, Anna and Ponti, Edoardo M. and Cohen, Shay B.},
  booktitle = {Advances in Neural Information Processing Systems},
  year      = {2024},
  doi       = {10.52202/079017-1815}
}

@misc{meta2025llama4scout,
  author       = {{Meta AI}},
  title        = {The {Llama} 4 Herd: The Beginning of a New Era of Natively Multimodal {AI} Innovation},
  year         = {2025},
  howpublished = {\url{https://ai.meta.com/blog/llama-4-multimodal-intelligence/}},
  url          = {https://ai.meta.com/blog/llama-4-multimodal-intelligence/}
}

@misc{qwen2025qwen3,
  title     = {{Qwen3} Technical Report},
  author    = {{Qwen Team}},
  year      = {2025},
  eprint        = {2505.09388},
  archivePrefix = {arXiv},
  url           = {https://arxiv.org/abs/2505.09388}
}

@misc{microsoft2024phi4,
  title     = {{Phi-4} Technical Report},
  author    = {Abdin, Marah and Aneja, Jyoti and Behl, Harkirat and Bubeck, S{\'e}bastien and Eldan, Ronen and Gunasekar, Suriya and Harrison, Michael and Hewett, Russell J. and Javaheripi, Mojan and Kauffmann, Piero and Lee, James R. and Lee, Yin Tat and Li, Yuanzhi and Liu, Weishung and Mendes, Caio C. T. and Nguyen, Anh and Price, Eric and de Rosa, Gustavo and Saarikivi, Olli and Salim, Adil and Shah, Shital and Wang, Xin and Ward, Rachel and Wu, Yue and Yu, Dingli and Zhang, Cyril and Zhang, Yi},
  year      = {2024},
  eprint        = {2412.08905},
  archivePrefix = {arXiv},
  url           = {https://arxiv.org/abs/2412.08905}
}

@misc{qwen2025qwen25vl,
  title     = {{Qwen2.5-VL} Technical Report},
  author    = {{Qwen Team}},
  year      = {2025},
  eprint        = {2502.13923},
  archivePrefix = {arXiv},
  url           = {https://arxiv.org/abs/2502.13923}
}

@misc{kim2024openvla,
  title     = {{OpenVLA}: An Open-Source Vision-Language-Action Model},
  author    = {Kim, {Moo Jin} and Pertsch, Karl and Karamcheti, Siddharth and Xiao, Ted and Balakrishna, Ashwin and Nair, Suraj and Rafailov, Rafael and Foster, Ethan and Lam, Grace and Sanketi, Pannag and Vuong, Quan and Kollar, Thomas and Burchfiel, Benjamin and Tedrake, Russ and Sadigh, Dorsa and Levine, Sergey and Liang, Percy and Finn, Chelsea},
  year      = {2024},
  eprint        = {2406.09246},
  archivePrefix = {arXiv},
  url           = {https://arxiv.org/abs/2406.09246}
}

@inproceedings{li2023camel,
  title     = {{CAMEL}: Communicative Agents for ``Mind'' Exploration of Large Language Model Society},
  author    = {Li, Guohao and Hammoud, Hasan Abed Al Kader and Itani, Hani and Khizbullin, Dmitrii and Ghanem, Bernard},
  booktitle = {Advances in Neural Information Processing Systems},
  year      = {2023},
  url       = {https://arxiv.org/abs/2303.17760},
  doi       = {10.52202/075280-2264}
}

@misc{wu2023autogen,
  title         = {{AutoGen}: Enabling Next-Gen {LLM} Applications via Multi-Agent Conversation},
  author        = {Wu, Qingyun and Bansal, Gagan and Zhang, Jieyu and Wu, Yiran and Li, Beibin and Zhu, Erkang and Jiang, Li and Zhang, Xiaoyun and Zhang, Shaokun and Liu, Jiale and Awadallah, Ahmed Hassan and White, Ryen W. and Burger, Doug and Wang, Chi},
  year          = {2023},
  eprint        = {2308.08155},
  archivePrefix = {arXiv},
  primaryClass  = {cs.AI},
  url           = {https://arxiv.org/abs/2308.08155}
}

@inproceedings{park2023generativeagents,
  title     = {Generative Agents: Interactive Simulacra of Human Behavior},
  author    = {Park, Joon Sung and O'Brien, Joseph C. and Cai, Carrie Jun and Morris, Meredith Ringel and Liang, Percy and Bernstein, Michael S.},
  booktitle = {Proceedings of the 36th Annual ACM Symposium on User Interface Software and Technology},
  year      = {2023},
  publisher = {Association for Computing Machinery},
  doi       = {10.1145/3586183.3606763},
  url       = {https://doi.org/10.1145/3586183.3606763}
}

@misc{hu2026evermembench,
      title={Evaluating Long-Horizon Memory for Multi-Party Collaborative Dialogues}, 
      author={Chuanrui Hu and Tong Li and Xingze Gao and Hongda Chen and Yi Bai and Dannong Xu and Tianwei Lin and Xiaohong Li and Yunyun Han and Jian Pei and Yafeng Deng},
      year={2026},
      eprint={2602.01313},
      archivePrefix={arXiv},
      primaryClass={cs.CL},
      url={https://arxiv.org/abs/2602.01313}, 
}

@misc{he2026memoryarena,
  title         = {{MemoryArena}: Benchmarking Agent Memory in Interdependent
                   Multi-Session Agentic Tasks},
  author        = {He, Zexue and Wang, Yu and Zhi, Churan and Hu, Yuanzhe
                   and Chen, Tzu-Ping and Yin, Lang and Chen, Ze
                   and Wu, Tong Arthur and Ouyang, Siru and Wang, Zihan
                   and Pei, Jiaxin and McAuley, Julian and Choi, Yejin
                   and Pentland, Alex},
  year          = {2026},
  eprint        = {2602.16313},
  archivePrefix = {arXiv},
  primaryClass  = {cs.CL},
  url           = {https://arxiv.org/abs/2602.16313}
}

@inproceedings{liu2025toolace,
  title     = {{ToolACE}: Winning the Points of {LLM} Function Calling},
  author    = {Liu, Weiwen and Huang, Xu and Zeng, Xingshan and Hao, Xinlong
               and Yu, Shuai and Li, Dexun and Wang, Shuai and Gan, Weinan
               and Liu, Zhengying and Yu, Yuanqing and Wang, Zezhong
               and Wang, Yuxian and Ning, Wu and Hou, Yutai and Wang, Bin
               and Wu, Chuhan and Wang, Xinzhi and Liu, Yong and Wang, Yasheng
               and Tang, Duyu and Tu, Dandan and Shang, Lifeng and Jiang, Xin
               and Tang, Ruiming and Lian, Defu and Liu, Qun and Chen, Enhong},
  booktitle = {The Thirteenth International Conference on Learning Representations},
  year      = {2025},
  url       = {https://openreview.net/forum?id=8EB8k6DdCU}
}

@misc{zou2023representation,
  title         = {Representation Engineering: A Top-Down Approach to {AI} Transparency},
  author        = {Zou, Andy and Phan, Long and Chen, Sarah and Campbell, James and Guo, Phillip and Ren, Richard and Pan, Alexander and Yin, Xuwang and Mazeika, Mantas and Dombrowski, Ann-Kathrin and Goel, Shashwat and Li, Nathaniel and Byun, Michael J. and Wang, Zifan and Mallen, Alex and Basart, Steven and Koyejo, Sanmi and Song, Dawn and Fredrikson, Matt and Kolter, J. Zico and Hendrycks, Dan},
  year          = {2023},
  eprint        = {2310.01405},
  archivePrefix = {arXiv},
  primaryClass  = {cs.LG},
  url           = {https://arxiv.org/abs/2310.01405}
}

@inproceedings{li2023iti,
  title     = {Inference-Time Intervention: Eliciting Truthful Answers from a Language Model},
  author    = {Li, Kenneth and Patel, Oam and Vi{\'e}gas, Fernanda and Pfister, Hanspeter and Wattenberg, Martin},
  booktitle = {Advances in Neural Information Processing Systems},
  volume    = {36},
  year      = {2023},
  doi       = {10.52202/075280-1797},
  url       = {https://proceedings.neurips.cc/paper_files/paper/2023/hash/81b8390039b7302c909cb769f8b6cd93-Abstract-Conference.html}
}

@misc{gemmateam2024gemma2,
  title         = {{Gemma 2}: Improving Open Language Models at a Practical Size},
  author        = {{Gemma Team}},
  year          = {2024},
  eprint        = {2408.00118},
  archivePrefix = {arXiv},
  primaryClass  = {cs.CL},
  url           = {https://arxiv.org/abs/2408.00118}
}

@misc{jiang2023mistral7b,
  title         = {{Mistral 7B}},
  author        = {Jiang, Albert Q. and Sablayrolles, Alexandre and Mensch, Arthur
                   and Bamford, Chris and Chaplot, Devendra Singh
                   and de las Casas, Diego and Bressand, Florian
                   and Lengyel, Gianna and Lample, Guillaume
                   and Renard Lavaud, L{\'e}lio and Saulnier, Lucile
                   and Lachaux, Marie-Anne and Stock, Pierre
                   and Le Scao, Teven and Lavril, Thibaut
                   and Wang, Thomas and Lacroix, Timoth{\'e}e
                   and El Sayed, William},
  year          = {2023},
  eprint        = {2310.06825},
  archivePrefix = {arXiv},
  primaryClass  = {cs.CL},
  url           = {https://arxiv.org/abs/2310.06825}
}

@misc{shekkizhar2025echoing,
  title         = {Echoing: Identity Failures when {LLM} Agents Talk to Each Other},
  author        = {Shekkizhar, Sarath and Cosentino, Romain and Earle, Adam and Savarese, Silvio},
  year          = {2025},
  eprint        = {2511.09710},
  archivePrefix = {arXiv},
  primaryClass  = {cs.AI},
  url           = {https://arxiv.org/abs/2511.09710}
}

@inproceedings{gupta2024taskinterference,
  title     = {{LLM} Task Interference: An Initial Study on the Impact of Task-Switch in Conversational History},
  author    = {Gupta, Akash and Sheth, Ivaxi and Raina, Vyas and Gales, Mark and Fritz, Mario},
  booktitle = {Proceedings of the 2024 Conference on Empirical Methods in Natural Language Processing},
  year      = {2024},
  url       = {https://aclanthology.org/2024.emnlp-main.811/},
  doi       = {10.18653/v1/2024.emnlp-main.811}
}

@misc{emde2026maseval,
  title         = {{MASEval}: Extending Multi-Agent Evaluation from Models to Systems},
  author        = {Emde, Cornelius and Rubinstein, Alexander and Goel, Anmol and Heakl, Ahmed and Yun, Sangdoo and Oh, Seong Joon and Gubri, Martin},
  year          = {2026},
  eprint        = {2603.08835},
  archivePrefix = {arXiv},
  primaryClass  = {cs.AI},
  url           = {https://arxiv.org/abs/2603.08835}
}

@misc{dongre2025driftnomore,
  title         = {Drift No More? Context Equilibria in Multi-Turn {LLM} Interactions},
  author        = {Dongre, Vardhan and Rossi, Ryan A. and Lai, Viet Dac and Yoon, David Seunghyun and Hakkani-T{\"u}r, Dilek and Bui, Trung},
  year          = {2025},
  eprint        = {2510.07777},
  archivePrefix = {arXiv},
  primaryClass  = {cs.CL},
  url           = {https://arxiv.org/abs/2510.07777}
}

@misc{qwen2024qwen25,
  title   = {{Qwen2.5} Technical Report},
  author  = {{Qwen Team}},
  year    = {2024},
  eprint        = {2412.15115},
  archivePrefix = {arXiv},
  url           = {https://arxiv.org/abs/2412.15115}
}

@misc{qwen2025qwen3vl,
  title   = {{Qwen3-VL} Technical Report},
  author  = {{Qwen Team}},
  year    = {2025},
  eprint        = {2511.21631},
  archivePrefix = {arXiv},
  url           = {https://arxiv.org/abs/2511.21631}
}
